\documentclass[lettersize,journal]{IEEEtran}
\usepackage{amsmath,amsfonts}
\usepackage{algorithmic}
\usepackage{bm}
\usepackage{algorithm}
\usepackage{array}
\usepackage{textcomp}
\usepackage{stfloats}
\usepackage{url}
\usepackage{verbatim}
\usepackage{graphicx}
\usepackage{cite}
\usepackage{xcolor}
\usepackage{lipsum}
\usepackage[font=small]{caption}
\usepackage{hyperref}
\usepackage{subfigure}
\usepackage{optidef}

\usepackage{siunitx}

\newcommand{\fig}[1]{Fig.~\ref{#1}}	
\newcommand{\eqn}[1]{Eq.~\eqref{#1}}	
\newcommand{\sect}[1]{Sec.~\ref{#1}}	

\newcommand{\R}[1]{\mathbb{R}^{#1}}
\newcommand{\N}{\mathbb{N}}
\newcommand{\SO}[1]{\mathsf{SO}(#1)}

\newcommand{\roll}{\phi}
\newcommand{\Lhook}{l_{\mathrm{hook}}}
\newcommand{\dhook}{d_{\mathrm{hook}}}
\newcommand{\dmin}{d_{\mathrm{min}}}
\newcommand{\dmax}{d_{\mathrm{max}}}
\newcommand{\dmaxh}{d_{\mathrm{max}}^{\mathrm{h}}}
\newcommand{\dmaxf}{d_{\mathrm{max}}^{\mathrm{t}}}
\newcommand{\hmin}{h_{\mathrm{min}}}
\newcommand{\phimax}{\phi_{\mathrm{max}}}

\newcommand{\wgrasp}{w_{\mathrm{gr}}}
\newcommand{\fgrasp}{f_{\mathrm{gr}}}
\newcommand{\ftension}{f_{\mathrm{t}}}
\newcommand{\plitter}{\mathbf{p}_{\mathrm{lt}}}
\newcommand{\tension}{t}
\newcommand{\g}{g}
\newcommand{\kgr}{k_{\mathrm{gr}}}
\newcommand{\NormTwo}[1]{\Vert {#1}\Vert_{2}}
\newcommand{\kpos}{k_{\mathrm{pos}}}
\newcommand{\Rz}{\mathbf{R}_{z}}

\newcommand{\arefr}{a^{\mathrm{ref}}_{\mathrm{rope}}}
\newcommand{\brefr}{b^{\mathrm{ref}}_{\mathrm{rope}}}
\newcommand{\psirefr}{\psi^{\mathrm{ref}}_{\mathrm{rope}}}
\newcommand{\drefr}{d^{\mathrm{ref}}_{\mathrm{rope}}}

\newcommand{\prefhook}{\mathbf{p}^{\mathrm{ref}}_{\mathrm{hook}}}
\newcommand{\xrefhook}{x^{\mathrm{ref}}_{\mathrm{hook}}}
\newcommand{\yrefhook}{y^{\mathrm{ref}}_{\mathrm{hook}}}
\newcommand{\zrefhook}{z^{\mathrm{ref}}_{\mathrm{hook}}}
\newcommand{\xdrefh}{\dot{x}^{\mathrm{ref}}_{\mathrm{hook}}}
\newcommand{\ydrefh}{\dot{y}^{\mathrm{ref}}_{\mathrm{hook}}}

\newcommand{\pplRA}{\mathbf{p}^{\mathrm{pl}}_{\mathrm{R}_1}}
\newcommand{\pplRB}{\mathbf{p}^{\mathrm{pl}}_{\mathrm{R}_2}}

\newcommand{\DeltaprefRB}{\Delta\mathbf{p}^{\mathrm{ref}}_{\mathrm{R}_2}}
\newcommand{\DeltaprefRA}{\Delta\mathbf{p}^{\mathrm{ref}}_{\mathrm{R}_1}}

\newcommand{\psiplRA}{\psi^{\mathrm{pl}}_{\mathrm{R}_1}}
\newcommand{\psiplRB}{\psi^{\mathrm{pl}}_{\mathrm{R}_2}}

\newcommand{\psirope}{\psi_{\mathrm{rope}}}
\newcommand{\phirope}{\phi_{\mathrm{rope}}}

\newcommand{\oW}{O_{\mathrm{W}}}

\newcommand{\oT}{O_{\mathrm{T}}}

\newcommand{\oVS}{O_{\mathrm{P}}}

\newcommand{\oRi}{O_{\mathrm{R}_i}}

\newcommand{\FW}{\mathcal{F}_{\mathrm{W}}}

\newcommand{\FT}{\mathcal{F}_{\mathrm{T}}}

\newcommand{\FVS}{\mathcal{F}_{\mathrm{P}}}
\newcommand{\FRA}{\mathcal{F}_{\mathrm{R}_1}}
\newcommand{\FRB}{\mathcal{F}_{\mathrm{R}_2}}
\newcommand{\FC}{\mathcal{F}_{\mathrm{C}}}
\newcommand{\FRi}{\mathcal{F}_{\mathrm{R}_i}}

\newcommand{\xW}{\mathbf{x}_{\mathrm{W}}}
\newcommand{\yW}{\mathbf{y}_{\mathrm{W}}}
\newcommand{\zW}{\mathbf{z}_{\mathrm{W}}}

\newcommand{\zB}{\mathbf{z}_{\mathrm{B}}}
\newcommand{\xT}{\mathbf{x}_{\mathrm{T}}}
\newcommand{\yT}{\mathbf{y}_{\mathrm{T}}}
\newcommand{\zT}{\mathbf{z}_{\mathrm{T}}}
\newcommand{\xRi}{\mathbf{x}_{\mathrm{R}_i}}
\newcommand{\yRi}{\mathbf{y}_{\mathrm{R}_i}}
\newcommand{\zRi}{\mathbf{z}_{\mathrm{R}_i}}

\newcommand{\yP}{\mathbf{y}_{\mathrm{R}}}

\newcommand{\posRA}{\mathbf{p}_{\mathrm{R}_1}}
\newcommand{\posRB}{\mathbf{p}_{\mathrm{R}_2}}

\newcommand{\Lrope}{l_{\mathrm{rope}}}
\newcommand{\mrope}{m_{\mathrm{rope}}}

\newcommand{\zatt}{z_{\mathrm{atc}}}

\newcommand{\wRrA}{{}^\mathrm{W}\mathbf{R}_{\mathrm{R}_1}}
\newcommand{\wRrB}{{}^\mathrm{W}\mathbf{R}_{\mathrm{R}_2}}

\newcommand{\Deltapvs}{\Delta\mathbf{p}^{\mathrm{vs}}}

\newcommand{\gfun}{g}

\newcommand{\spab}{\mathbf{s}}
\newcommand{\avs}{a_{\mathrm{p}}}
\newcommand{\bvs}{b_{\mathrm{p}}}
\newcommand{\cvs}{c_{\mathrm{p}}}
\newcommand{\phivs}{\phi_{\mathrm{p}}}
\newcommand{\psivs}{\psi_{\mathrm{p}}}

\newcommand{\sref}{\mathbf{s}^{\mathrm{ref}}}
\newcommand{\avsref}{a_{\mathrm{p}}^{\mathrm{ref}}}
\newcommand{\bvsref}{b_{\mathrm{p}}^{\mathrm{ref}}}
\newcommand{\psivsref}{\psi_{\mathrm{p}}^{\mathrm{ref}}}

\newcommand{\vrel}{\mathbf{v}_{\mathrm{rel}}^{\mathrm{vs}}}
\newcommand{\relDis}{\Delta\mathbf{p}^{\mathrm{vs}}}
\newcommand{\kc}{k_c}
\newcommand{\ki}{k_i}
\newcommand{\es}{\mathbf{e}_{\mathrm{s}}}
\newcommand{\esi}{\mathbf{e}_{\mathrm{s},i}}

\newcommand{\Mmat}{\mathbf{M}}

\newcommand{\PW}{\mathrm{P}_{\mathrm{W}}}
\newcommand{\PT}{\mathrm{P}_{\mathrm{P}}}
\newcommand{\pattA}{\mathbf{p}_{\mathrm{A}_1}}
\newcommand{\pattB}{\mathbf{p}_{\mathrm{A}_2}}

\newcommand{\pIn}{\mathbf{p}_{i}}

\newcommand{\np}{n_p}
\newcommand{\nvec}{\mathbf{n}}
\newcommand{\nvecT}{\mathbf{n}^\top}

\newcommand{\zeroThree}{\mathbf{0}_3}
\newcommand{\pattBT}{{}^{\mathrm{P}}\mathbf{p}_{\mathrm{A}_2}}
\newcommand{\pUnoT}{{}^{\mathrm{P}}\mathbf{p}_{\mathrm{1}}}
\newcommand{\pInT}{{}^{\mathrm{P}}\mathbf{p}_{i}}
\newcommand{\pNT}{{}^{\mathrm{P}}\mathbf{p}_{\np}}

\newcommand{\wgps}{w_{\mathrm{p}}}
\newcommand{\wmeas}{w_{\mathrm{m}}}
\newcommand{\wdepth}{w_{\mathrm{d}}}

\newcommand{\xAtB}{x_{\mathrm{A}_2}}
\newcommand{\yAtB}{y_{\mathrm{A}_2}}
\newcommand{\zAtB}{z_{\mathrm{A}_2}}

\newcommand{\arcsinh}{\text{arcsinh}}

\begin{document}

\title{Multi-robot Aerial Soft Manipulator \\ For Floating Litter Collection}

\author{Antonio González-Morgado$^1$, Sander Smits$^2$, Guillermo Heredia$^1$,
Anibal Ollero$^1$,\\ Alexandre Krupa$^3$, François Chaumette$^3$, Fabien Spindler$^3$, 
  Antonio Franchi$^{2,4}$, and Chiara Gabellieri$^2$
\thanks{Corresponding author: A. Gonzalez-Morgado (\url{mantonio@us.es}).}
\thanks{$^1$ GRVC Robotics Lab, Escuela Tecnica Superior de Ingenieria,
Universidad de Sevilla, 41012 Sevilla, Spain.} 
\thanks{$^2$Robotics and Mechatronics Department, Electrical Engineering,  Mathematics, and Computer Science (EEMCS) Faculty, University of Twente, 7500 AE Enschede, The Netherlands.}
\thanks{$^3$Inria, Univ. Rennes, CNRS, IRISA, 35042
Rennes, France.}
\thanks{$^4$Department of Computer, Control and Management Engineering, Sapienza University of Rome, 00185 Rome, Italy.}
\thanks{This work was supported in part by the European Commission through the AEROTRAIN Marie Skłodowska-Curie project under Grant MSCA-ITN-2020-953454 and through the Flyflic Marie Skłodowska-Curie project under Grant MSCA-PF 101059875.}
}



\maketitle

\begin{abstract}

Removing floating litter from water bodies is crucial to preserving aquatic ecosystems and preventing environmental pollution. In this work, we present a multi-robot aerial soft manipulator for floating litter collection, leveraging the capabilities of aerial robots. The proposed system consists of two aerial robots connected by a flexible rope manipulator, which collects floating litter using a hook-based tool. Compared to single-aerial-robot solutions, the use of two aerial robots increases payload capacity and flight endurance while reducing the downwash effect at the manipulation point, located at the midpoint of the rope. Additionally, we employ an optimization-based rope-shape planner to compute the desired rope shape. The planner incorporates an adaptive behavior that maximizes grasping capabilities near the litter while minimizing rope tension when farther away. The computed rope shape trajectory is controlled by a shape visual servoing controller, which approximates the rope as a parabola. The complete system is validated in outdoor experiments, demonstrating successful grasping operations. An ablation study highlights how the planner's adaptive mechanism improves the success rate of the operation. Furthermore, real-world tests in a water channel confirm the effectiveness of our system in floating litter collection. These results demonstrate the potential of aerial robots for autonomous litter removal in aquatic environments.

\end{abstract}



\section{Introduction}
In 2019, 353 million tonnes of plastic waste was generated, only  9\% of which was recycled, while 22\% was mismanaged, with a considerable portion of it ending up in water. The best solutions to plastic pollution include preventing it from entering the environment, e.g., by limiting single-use plastic, and improving plastic management \cite{horejs2020solutions}. However, the OECD projects that the volume of mismanaged plastic will still doubled by 2060 compared to now,  despite better recycling rates \cite{sonke2024global}.

Rivers have been identified as one of the main ways through which plastic reaches the oceans \cite{helinski2021ridding}, although exact estimates of the rivers' contribution to plastic input in the oceans are diverging \cite{horejs2020solutions}. The degradation of macroplastic in water leads to the accumulation of microplastic, the absorption of which causes adverse effects on aquatic organisms by threatening biodiversity, harming individual organisms, affecting mobility, survival, reproduction, to name a few \cite{koelmans2022risk}. Although clear assessment of the risk posed by such exposure on humans is mostly lacking, microplastics have already been found in human organs \cite{kozlov2025your}. 

Together with reducing the amount of plastic entering the ocean and preventing plastic from harming riverine life, clearing rivers from debris has also other crucial societal benefits, such as reducing the flood risks \cite{honingh2020urban}. 

Manual collection of floating litter by human operators is a demanding and potentially hazardous solution. Slippery branches, not-easy-to-reach spots, and harsh environmental conditions are sources of risks to humans. Autonomous solutions to the collection of floating litter have been proposed both in the scientific literature and on the market. 

Autonomous solutions can be mostly grouped into two main categories: fixed mechanisms and autonomous or semi-autonomous boat-like solutions. Trapping mechanisms such as 
those described in~\cite{rivercleaning,seads, watergoat, oceancleanup} belong to the former category. They are typically composed of floating elements creating a physical impediment to the passage of debris and, thus, accumulating them. The accumulated litter requires to be then collected, for instance by a separate system, such as the floating conveyor belt proposed by \cite{oceancleanup} or the integrated solution proposed by \cite{noria}, where the litter is conveyed to a collection gear on the side of the river thanks to the funnel shape of the trapping mechanism. While floating barriers are the most common solutions in this group, there are other systems, such as \cite{lee2024optimal, seabin}, designed to actively suck rubbish even in locations where there is no natural water stream, see, e.g., marinas, ports, ponds, etc. With the other solutions, this shares the need to be periodically emptied. 
In general, fixed mechanisms may be very suitable to address large quantities of litter in strategic spots, e.g., the delta of huge rivers, but have the drawback of lacking capillarity and flexibility. Moreover, some of them also hinder navigation, so their applicability is limited to restricted areas. \cite{bubblebarrier} creates a vertical layer of bubble thanks to a perforated tube placed on the bottom of the waterway where air is pumped through. While it does not impede navigability, it shares with other fixed solutions the lack of flexibility.

On the other hand, boat-like autonomous solutions like~\cite{10165589,zhu2022smurf, seavax, wasteshark,jellyfishbot} can target specific spots at need, allowing a more capillary collection of floating waste. However, they can only access navigable waterways, they typically have a considerable cost \cite{helinski2021ridding}, and their deployment inside and outside of the water is not trivial. Such a drawback is shared by amphibious robotic solutions \cite{zhang2023spiral,muthusamy2024novel} designed for land- and water-garbage collection.

Aerial robots have been proposed in the past as passive tools to \textit{monitor} riverine plastic litter \cite{geraeds2019riverine}. However, they have the potential to actively tackle the problem, too. Their large workspace allows them to skip dams and non-navigable spots such as small falls, interruptions where the canals go underwater, and frozen or dry portions, etc; they are easily deployed in and out of the manipulation site, and they prevent humans from the need to get closer to slippery branches to accomplish the task manually. Hence, they have the potential to represent a flexible and low-cost solution to the addressed challenge and complement the action of other proposed solutions. In \cite{zoric2024towards}, an aerial robotic prototype for floating litter collection was proposed for the first time. One quadrotor equipped with a suspended net was tested indoors. Data-driven methods were integrated for litter segmentation and a primitive-based trajectory was planned for the net.
A multi-robot approach, where two UAVs uses a suspended cable, has been theorized in \cite{gabellieri2023differential} for increasing the overall payload and separating the robot downwash from the manipulation site. 

The cable has the advantage of being lightweight and inexpensive. However, the control of flexible cables hanging below aerial robots is a timely challenge and is recognized to have many potential applications in, e.g., manipulation of fluid conduits \cite{abhishek2021towards} and fire fighting \cite{kotaru2020multiple} besides fresh waterways 
management. The work in~\cite{kotaru2020multiple} shows the differential flatness of single- and multi-robot systems connected to flexible cables modeled as interconnected masses and rigid links, and it proposes a controller over the linearized system.    A step forward is provided by~\cite{gabellieri2023differential} which demonstrates the differential flatness of a discrete cable model that embeds elasticity. Such discretized models have the drawback of increased complexity when the cable is finely decomposed into a large number of segments.   A different modeling approach is proposed  in~\cite{d2021catenary, abhishek2021towards, abiko2017obstacle}, where two quadrotors are connected by a cable modeled as a catenary curve. Such a method allows describing the shape using a reduced set of parameters and has been tested in indoor experiments~\cite{d2021catenary, abiko2017obstacle}. In order to achieve lightweight computations and robustness to singularities, the parabola has been proposed to model the cable hanging between two aerial robots as an alternative to the catenary curve~\cite{estevez2022hybrid}. In~\cite{smolentsev2023shape}, a 
visual servoing method using a RGB-D camera and such a parabola model was introduced to enable a robotic arm to autonomously manipulate a flexible cable under the influence of gravity, guiding it toward a desired shape. This approach was later implemented in~\cite{smolentsev2024shape} to control a suspended flexible cable attached between two quadrotor drones in indoor scenarios.

Motivated by the higher payload capacity and extended flight time of multi-robot aerial systems, as well as the flexibility provided by ropes, this work presents a multi-robot aerial soft manipulator for floating litter collection. The proposed system consists of two aerial robots connected by a flexible rope, which serves as a manipulator for floating litter retrieval. Due to the high number of possible shape configurations of the flexible manipulator, we introduce a rope-shape planner that computes the optimal rope shape for floating litter collection. This planner maximizes grasping capabilities when the manipulator is near the litter while minimizing rope tension when it is farther away. To control the rope shape, we integrate a visual servoing controller into a fully functional robotic prototype. To address this, the visual servoing method from \cite{smolentsev2023shape,smolentsev2024shape} is first tested in outdoor environments, where changing light conditions and reflections on the water surface pose challenges for vision-based sensors. Based on these first results, we propose a modified version of the visual servoing controller from \cite{smolentsev2023shape,smolentsev2024shape}, where control actions were previously applied to only one robot, distinguishing between leader and follower roles. Instead, we symmetrically distribute the visual servoing control action between both robots. This approach enables precise control of the rope shape without altering the planned trajectory of the middle rope point, where the end-effector for litter collection is installed. 
Additionally, as for the visual perception of the rope, we refine the estimation of the parabola parameters by formulating the problem as an optimization task, where different weights regulate its behavior to account for sensor uncertainty when generating interpolated data. The complete system is extensively validated through outdoor experiments. First, validation experiments are conducted in a grass field to test the system under controlled conditions. Then, an ablation study, performed for different trajectories and optimization weights, highlights the importance of the adaptive behavior of the planner in increasing grasping success. Finally, additional experiments in a real water channel confirm the applicability and effectiveness of our approach for floating litter collection.

\begin{figure*}[b]
	\centering
	\includegraphics[width=2\columnwidth]{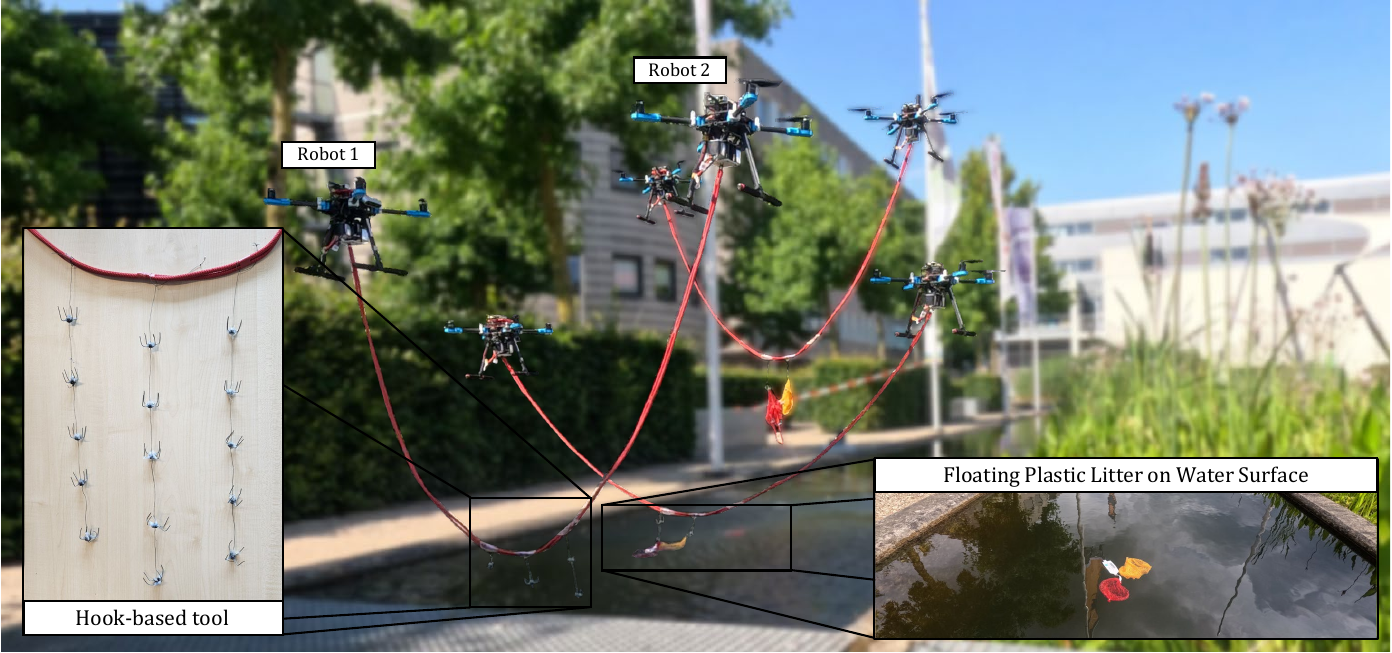}
	\caption{Floating litter collection with an aerial soft manipulator system. Two aerial robots, robot 1 and robot 2, control the shape of a flexible rope to collect floating litter in a water channel. The rope includes a hook-based tool for grasping the floating litter.}
	\label{fig:task_def}
\end{figure*}

\section{Results}
\subsection{Task definition}

In this work, we study the task of collecting floating litter from water bodies such as channels or lakes, as shown in \fig{fig:task_def}. To achieve this, we propose the use of an aerial system composed of two aerial robots, robot 1 and robot 2, that control the shape of a flexible rope equipped with a tool for collecting floating litter from the water.

\fig{fig:task_def} illustrates sequential snapshots of the task performed in a real-world water channel. The operation begins with a manual takeoff, after which both drones autonomously follow a coordinated trajectory, while actively controlling the shape of the rope. The desired rope shape is computed offline by an optimal rope-shape planner, which tracks a predefined trajectory of the rope’s lowest point to collect litter from the water surface using the installed tool. The planner generates trajectories for both rope endpoints attached to the robots and for the rope shape, which is parameterized as a parabola. The endpoint trajectories are provided to the robots as nominal references. However, to enhance robustness against disturbances, both robots adapt their planned trajectories to ensure the rope maintains the desired parabolic shape commanded by the planner. To achieve this, one of the robots, refereed as robot 1, is equipped with a depth camera that estimates the rope’s shape. A shape visual servoing controller is employed to control the rope’s shape using feedback from the onboard camera. The computed control action modifies both planned trajectories.

In the setup chosen for the tests, the flexible rope includes a hook-based tool designed to collect litter from the water surface, as shown on the left side of \fig{fig:task_def}. This end-effector is suitable for trapping floating plastic bags and similar items, as they get trapped by the hooks. However, the selected end-effector does not limit the applicability of our method for floating litter collection, as other types of floating litter, such as cans or plastic bottles, can be captured using alternative end-effectors, such as nets. The exploration of different end-effectors is beyond the scope of this work. 


We assume that the litter positions are known for the offline planner, which is reasonable under the hypothesis of still waters. Additionally, by using two drones positioned away from the grasping point of the rope, the disturbances caused by the propeller airflow are reduced, preventing disruption of the litter’s position during the execution of the task. In practice, the litter location can be retrieved online using existing trash detection and tracking solutions \cite{zoric2024towards,geraeds2019riverine,gutsa2024wasted,gallitelli2024monitoring}. Once the collection rope passes over the litter, both drones ascend to lift the litter from the water, trapping it on the hooks of the flexible rope. Finally, the drones land, allowing an operator to retrieve the collected litter from the hooks, leaving the system ready for the next operation.

\subsection{Design rationale}

The proposed aerial manipulator system is based on the use of aerial robots, which offer a large workspace and the ability to access hard-to-reach areas. The manipulator consists of a flexible, lightweight, and inexpensive rope, connected at both ends to two drones. 
The use of a rope-based manipulator enhances adaptability compared to rigid alternatives, as its flexibility reduces the impact of disturbances on the end-effector compared to rigid designs~\cite{d2021catenary}. In this configuration, the shape of the rope is controlled by modifying the position of both ends via the drones.

A key advantage of using two drones to control the rope shape is the significant reduction in downwash in the manipulation area, which coincides with the midpoint of the rope, where the end-effector is installed. This reduces disturbances on the litter, preventing its displacement during the operation and ensuring it remains in the same position. Additionally, employing two drones enhances payload capacity or flight time compared to single-drone-based solutions~\cite{meng2020survey,ollero2021past}.

Furthermore, the flexible manipulator is equipped with a lightweight end-effector composed of thin threads with hooks, as shown on the left side of \fig{fig:task_def} (manufacturing details are available in the Methods section). This design enables the collection of floating litter simply by dragging the hooks across the water surface. The hooks capture the litter, allowing it to be retrieved by an operator after landing. Thanks to this configuration only the hooks make contact with the water. This minimizes drag on the rope manipulator, as there is no direct water interaction, and ensures that the rope remains fully visible to the depth camera for accurate shape estimation.

\subsection{Rope-shape planning rationale}

\begin{figure*}[t]
	\centering
	\includegraphics[width=2\columnwidth]{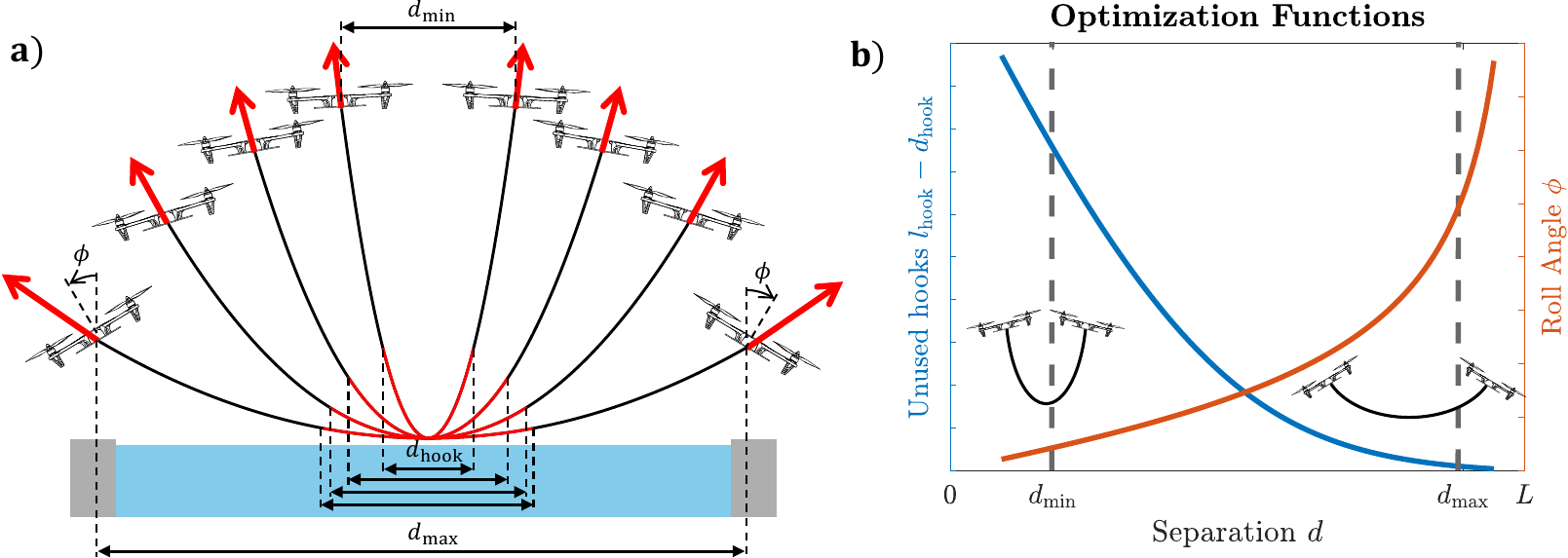}
	\caption{Rope shape for litter collection. a) Different rope shapes during the litter collection, including the needed force (red arrows) and the width covered by the hook tool (red segments in the middle zone of the rope). b)  Representation of the behavior of the two optimization functions, including the needed roll angle of the UAVs ($\roll$) and the unused hook width ($\Lhook-\dhook$). 
    }
	\label{fig:rope_shape}
\end{figure*}

The desired configuration of the rope during the task is determined by an offline rope-shape planner. This planner takes as input a predefined reference trajectory $\prefhook$ for the middle point of the rope, where the hook tool is installed. This trajectory is assumed to be provided by the user and can be generated using state-of-the-art techniques for computing smooth paths over waypoints \cite{richter2016polynomial}. The output of the planner is the planned trajectories of the rope extremes, $\pplRA$ and $\pplRB$, controlled by the drones, and the desired shape of the rope, parametrized as a parabola of parameters $\arefr$ and $\brefr$, where the curve is given by $z=\arefr y^2+\brefr y$ when it is expressed in the rope frame.

Since the end-effector is installed at the middle point of the rope, 
this point should be positioned as close as possible to the water surface to facilitate the grasping of litter, while the remaining segments of the rope stay above the water. To achieve this, we impose a symmetric parabolic shape ($\brefr(t) =0$), by ensuring that the two robots follow symmetric flight paths with respect to the parabola’s center. By doing this, the shape of this parabola is determined by only two main parameters: the horizontal separation $\drefr$ of the drones, which characterizes the curvature $\arefr$ of the profile, and the yaw angle $\psirefr$ of the parabola, which defines its orientation in space. 

The yaw angle $\psirefr$ is selected to maintain a perpendicular orientation relative to the desired trajectory $\prefhook$ of the middle point. This alignment optimizes the grasping capability of the hooks by maximizing the effective contact surface with the litter. The horizontal separation $\drefr$ of the parabola's extremes is obtained by solving an optimization problem that minimizes two objectives: (1) the overall tension in the rope, which must be compensated by the robots, and (2) the influence of positioning errors at the middle point, which directly affects the effectiveness of the litter collection process. In the following, we provide a detailed analysis of these optimization criteria and their impact on the overall performance of the system. The mathematical formulation and implementation details are presented in the Methods section.

As already said just above, to generate the desired parabolic shape of the rope, the optimization problem includes two objective terms. The first objective aims to minimize the tension in the rope, which must be compensated for by both drones at the rope's extremities. This tension results from the need to counteract the total weight of the rope and the grasped object. The vertical component of the force exerted by each drone compensates for this weight, while the horizontal components counterbalance each other at opposite ends. The lateral force is generated by the drones by adjusting their roll angle $\phi$; therefore, the greater the required lateral force, the larger the required roll angle. As the total force at each extreme is tangent to the rope’s shape at that point, reducing the distance between drones decreases the horizontal force component, and then the required roll angle, as shown in \fig{fig:rope_shape}a. Ideally, both drones would be positioned at the same point, leading to a purely vertical rope shape and eliminating lateral forces. However, this is not feasible, as a minimum separation distance is required for safe operation. To enforce this constraint, the optimization problem includes an inequality constraint that guarantees a minimum distance $\dmin$ between the drones. The second objective term aims to minimize the sensitivity of the grasping operation to positioning errors in the rope. To achieve this, we maximize the hook’s covered width $\dhook$ on the water, ensuring that the hooks effectively grasp litter. Similarly, this is accomplished by minimizing the unused hook width, represented by $\Lhook - \dhook$, where $\Lhook$ is the length of the rope along which the hooks are deployed (see the red rope length in \fig{fig:rope_shape}a). The rope section responsible for grasping must be close enough to the water surface for the hooks to maintain contact. When the lowest point of the rope reaches the water, increasing the drone separation extends the portion of the rope that can grasp litter, represented by $\dhook$ in \fig{fig:rope_shape}a. However, excessive separation increases tension in the rope, potentially leading to unfeasible force demands on the drones. Additionally, as the drones move further apart, they must fly closer to the water to maintain the middle point at an appropriate height for litter collection. To address these trade-offs, the optimization problem includes an upper bound $\dmax$ on the drone separation distance, considering the maximum lateral force each platform can generate and a minimal height of flight. Figs.~\ref{fig:rope_shape}a
and~\ref{fig:rope_shape}b provide an interpretation of the objective functions and the maximum and minimal distance between drones. 

\begin{figure*}[t]
	\centering
	\includegraphics[width=2\columnwidth]{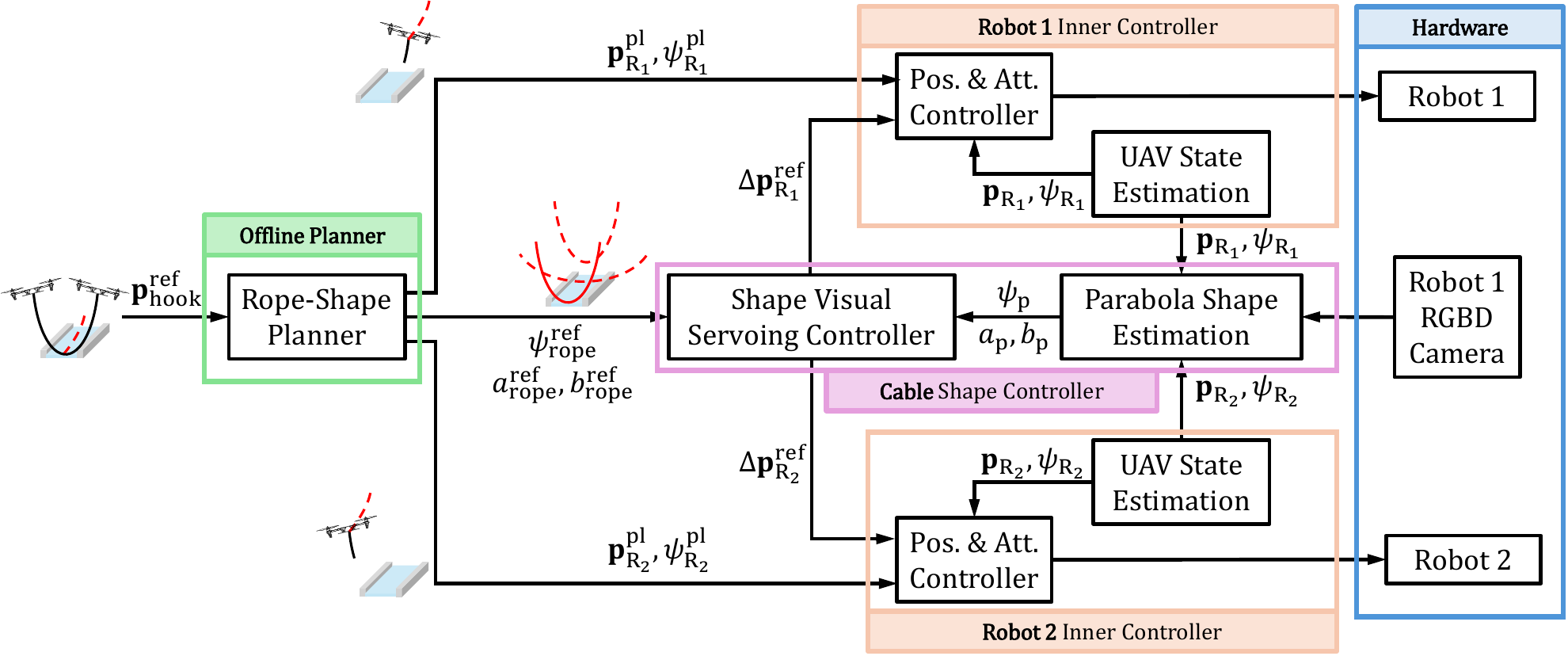}
	\caption{Control framework for litter collection, composed by the offline rope-shape planner (green box), the rope shape controller (pink box) and the inner controllers for the aerial robots (orange boxes). The offline planner receives as input the hook reference trajectory ($\prefhook$), and generates the reference trajectory of robot 1 ($\pplRA,\psiplRA$), the reference trajectory of robot 2 ($\pplRB,\psiplRB$) and the rope shape trajectory ($\arefr,\psirefr$). }
	\label{fig:control_sh}
\end{figure*}

As the two objective functions have opposite behaviors—when the drones are more separated, the unused hook length is reduced, but the required roll angle to compensate the lateral forces increases—the final behavior is controlled through a weighting factor. Furthermore, since minimizing the unused hook length objective function is only relevant when the rope is close to the litter’s position, its weight is adapted dynamically based on the distance to the litter. This adaptive weighting strategy ensures that rope tension is continuously minimized while prioritizing grasping performance only when the rope approaches the litter. 

Finally, the desired parabola shape, defined by its yaw angle $\psirefr$ and the desired separation between drones $\drefr$, is combined with the rope’s middle point trajectory $\prefhook$ to generate the trajectories of the aerial robots. These trajectories are commanded as planned references $\pplRA,\pplRB$ for both robots. However, these trajectories can be modified to control the planned parabola shape, enhancing robustness in maintaining the desired configuration of the rope. Furthermore, since the rope is connected to the drones via a spherical joint, they can perform yaw rotations without affecting the rope’s shape. As a result, the yaw angles of both drones are commanded to match the rope’s yaw angle. This ensures that the camera mounted on robot 1 maintains optimal visibility of the rope, keeping it centered in the image for accurate shape estimation. In the following section, the complete control framework for controlling the drones’ trajectories and the rope’s shape is presented.

\subsection{Control Rationale}

The proposed control framework enables the aerial robots to track the planned rope shape $\arefr,\brefr$ and $\psirefr$. It consists of two main components: (1) a shape visual servoing controller, which controls the shape of the parabola during the operation, and (2) an inner controller for each drones, responsible for tracking the corrected trajectory of the drones. The mathematical formulation and implementation details are described in the Methods section. \fig{fig:control_sh} shows the overall control scheme, including the offline rope-shape planner presented before.

The shape visual servoing controller adjusts the nominal trajectory of the drones to ensure that the rope maintains the planned shape, while the inner controller precisely follows this corrected trajectory. To achieve this, the correction terms $\DeltaprefRA,\DeltaprefRB$ in the nominal trajectory incorporates the computed error in the parabola parameters, defined as the difference between the planned and current parabola parameters. The control law implements a Proportional-Integral (PI) controller, similar to \cite{smolentsev2024shape}, where the parameter errors are transformed into linear velocity corrections using the interaction matrix (see Methods section for more details). The control action of this visual servoing controller is a desired relative displacement $\relDis$ between both robots. In \cite{smolentsev2024shape}, this correction term $\relDis$ is directly applied to robot 1, while robot 2 is not modified. In our application, this approach would modify the final trajectory of the end-effector. To avoid this, we applied the correction term in a symmetrical way to both robots ($\DeltaprefRA=-\relDis/2$ and $\DeltaprefRB=\relDis/2$), which guarantees that the hook follows the planned trajectory. 
Finally, these corrections are incorporated into the reference trajectory of the robots, as shown in \fig{fig:control_sh}

The parabola parameters are estimated using the data from the onboard RGB-D camera. The proposed pipeline is based on \cite{smolentsev2024shape}, incorporating some improvements. The pipeline starts with the data acquisition from the camera, capturing both the point cloud and the RGB image. The rope is identified in the RGB image by segmenting its red color, followed by binarization to isolate the rope’s points in the depth data. These points, initially in the camera frame, are then transformed into the world frame using the current position of robot 1 and the depth data.

At this stage, we refine the algorithm from \cite{smolentsev2024shape} to enhance the robustness of the estimation. The current position of both drones is also incorporated into the rope’s point cloud, as they represent its endpoints. The yawing direction of the parabola $\psirope$ is then estimated by determining the normal direction of the best-fitting plane for the generated point cloud. Next, the parabola parameters are obtained by solving an optimization problem that minimizes the estimation error for the generated point cloud. Since the point cloud is constructed from multiple sources—primarily the depth camera, along with the two drone positions given by the GPS—the weighting factors in the optimization problem accounts for the varying levels of noise in these data sources. To further enhance robustness, we introduce a constraint on the total rope length, ensuring a feasible and accurate estimation even in cases where the rope is not identified at all by 
the camera and only the two endpoints are available. The output of the optimization process consists of the estimated parabola parameters. To reduce noise, a standard Kalman filter is applied to the estimated cable plane parameters. Finally, the estimated parabola parameters are used by the visual servoing controller to regulate the shape of the rope.

The inner controller of the drones is designed to follow the commanded trajectories, which are generated by the shape planner and corrected by the visual servoing controller. In our case, we directly utilize the control architecture implemented in the open-source autopilot PX4. This autopilot employs PID-cascaded structures for both attitude and position control, with implementation details provided in Supplementary Method 1.

\subsection{Experimental validation}

\begin{figure*}[b]
	\centering
	\includegraphics[width=2\columnwidth]{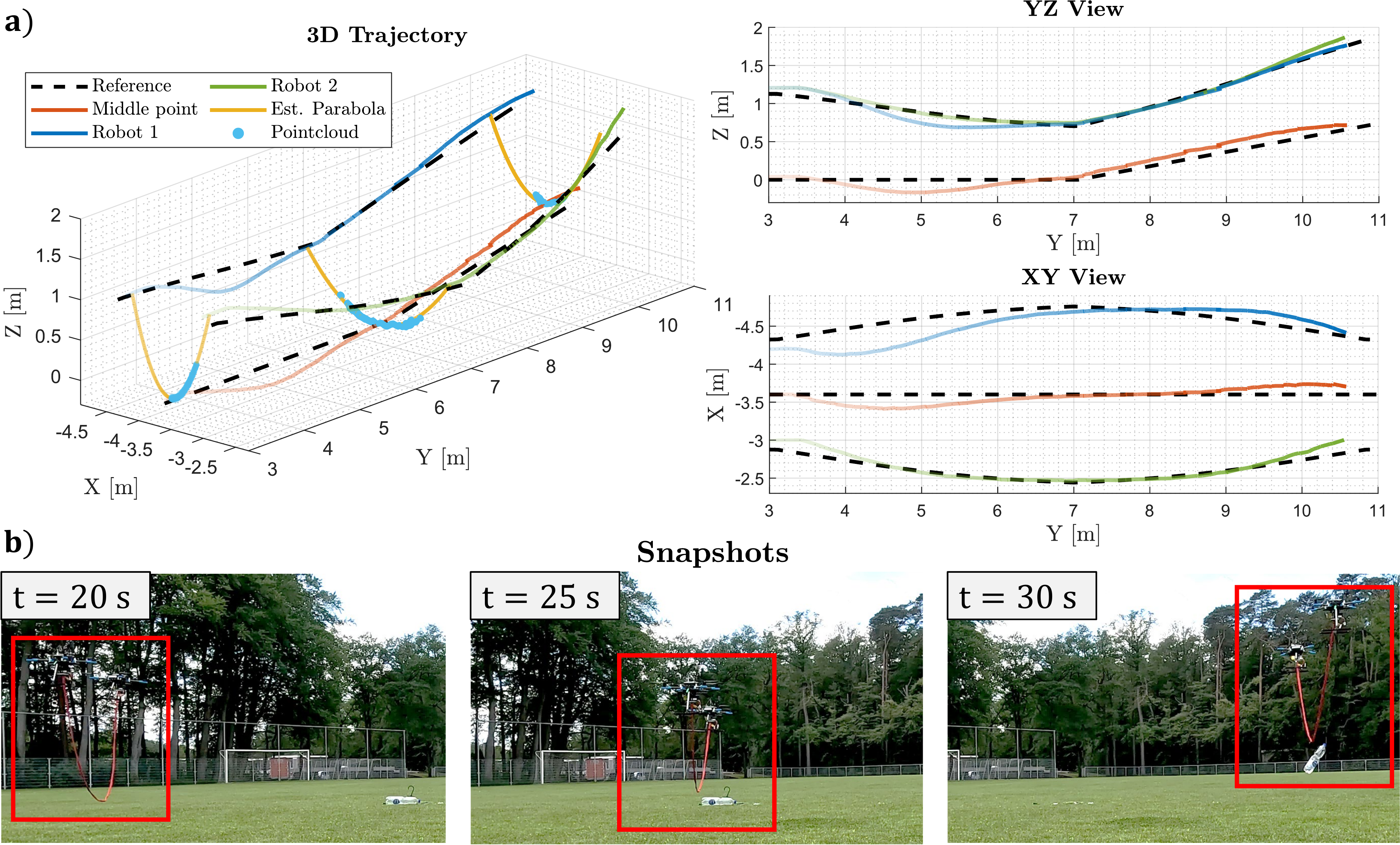}
    \caption{Results of the experiments in the grass field. a) 3D and 2D views of the trajectories. The transparency effect indicates the temporal progression of the trajectories, transitioning from transparent (initial) to solid (final) lines. The trajectories of robot 1 (blue line), robot 2 (green line), and the rope's middle point (red line) are represented as solid lines, while the reference trajectories are represented as black dashed lines. In addition, the estimated parabola shape (yellow lines) and the pointcloud generated by the depth camera are represented for three different time instants. b) Snapshots of the validation experiment for three different time instants.}\label{fig:ValExp_Traj}
\end{figure*}

The complete aerial manipulator system is validated through outdoor experiments. First, the system is validated in a grass field where the rope manipulator, without the hook-based tool, has to grasp an object with the rope. Then, an ablation study is performed to evaluate the influence of the weights of the optimal planner in the success rate of the grasping operation. Finally, the system is validated in a water channel collection floating litters with the hook based tool.

\subsubsection{Grass field validation experiment} The first experiment aims to validate and analyze the system under controlled conditions. To achieve this, the aerial manipulator with the rope flights over a grass field, where an object with an attached hook must be grasped by the rope. In our case, we use a plastic bottle with a 3D-printed hook as an object to be grasped. The bottle included a small amount of water, resulting in a total mass of approximately $150\ \si{\gram}$, to prevent possible displacement caused by downwash, especially under non-optimal rope configurations during the ablation study.
This setup allows precise control of the litter's position, preventing external factors such as wind, downwash, or currents from altering its location, as would happen in water. The wind during this experiment was $3\ \si{\meter\per\second}$.  The experimental setup is detailed in the Methods section, while the video of the experiment is included in Supplementary Movie S1.

\begin{figure*}[t]
	\centering
	\includegraphics[width=2\columnwidth]{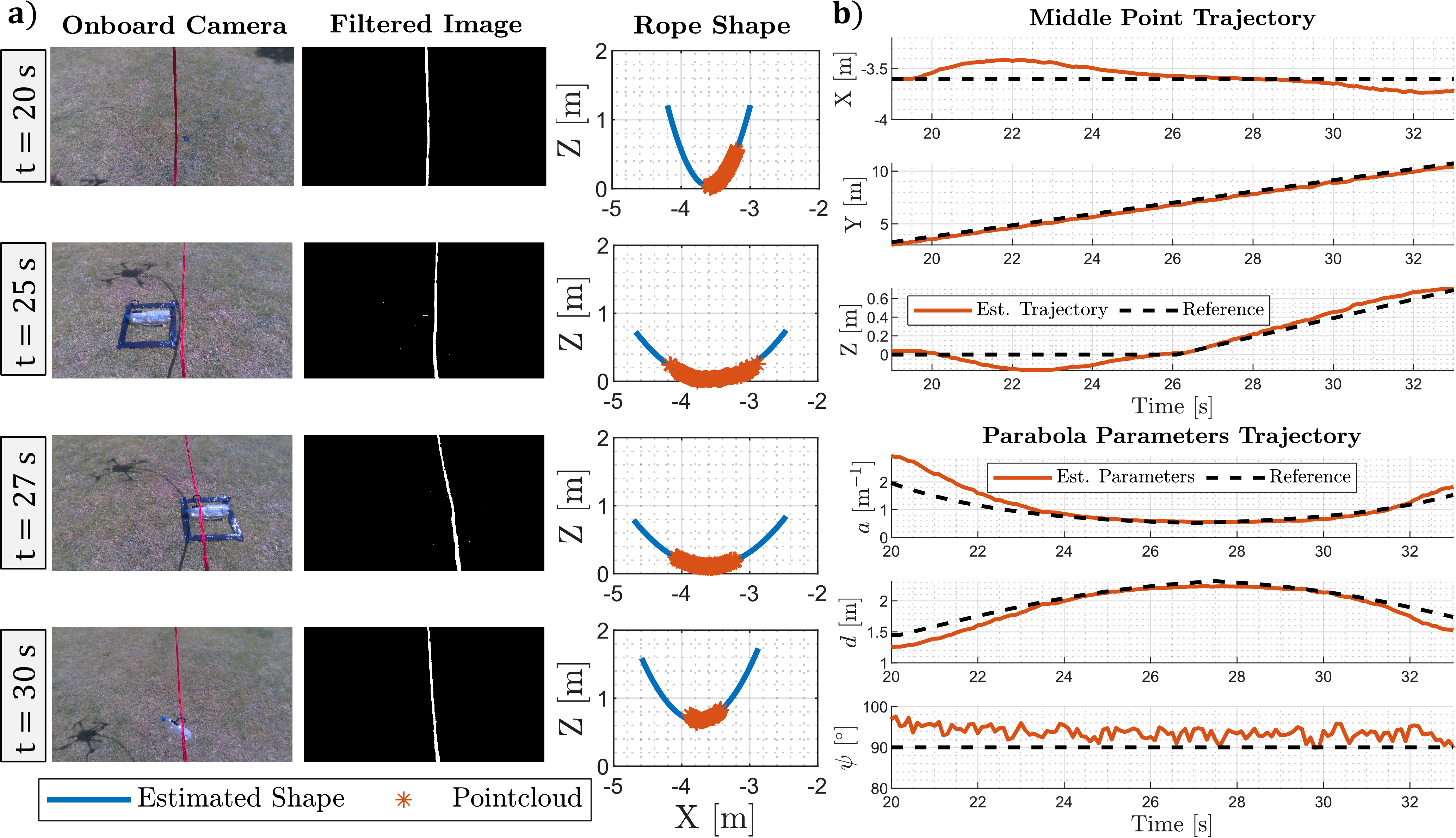}
    \caption{Parabola shape estimation during the validation experiment in the grass field. a) Onboard camera view (first column), filtered image (second column) and filtered pointcloud and estimated parabola shape (third column). The four time instants corresponds with $t=20\ \si{\second}$ (first row), $t=25\ \si{\second}$ (second row), $t=27\ \si{\second}$ (third row), and $t=30\ \si{\second}$ (fourth row).  b) Rope trajectories, including the middle point trajectory (top) 
    and the parabola parameter trajectory (bottom). The shape parameters are the curvature $a$, the separation $d$ and the yaw angle $\psi$. 
    }
    \label{fig:ValExp_cam}
\end{figure*}

In this case, the litter to grasp is located at $\plitter=[-3.6,7.0,0.0]^\top\si{\meter}$. The initial position of the rope middle point is $[-3.6,3.0,0.0]^\top\si{\meter}$. For collecting the litter, the middle point needs to follow a straight trajectory in the $\yW$ axis until reaching the litter position $\plitter$, with a constant velocity of $0.5\ \si{\meter\per\second}$ in the $\yW$ axis. After reaching the liter position $\plitter$, the middle rope point ascent at a constant velocity of $0.1\ \si{\meter\per\second}$, while moving forward in the $\yW$ axis at $0.5\ \si{\meter\per\second}$. \fig{fig:ValExp_Traj}a and \fig{fig:ValExp_cam}b-top show the reference trajectory of the middle point. 

With the middle-point reference trajectory $\prefhook$ defined by the user, and the litter position $\plitter$, the rope shape planner computes the trajectories of the robots and the rope shape. \fig{fig:ValExp_Traj}a shows the robots' reference trajectories, while \fig{fig:ValExp_cam}b-bottom presents the rope reference parameters. Notably, the reference separation increases when approaching the litter's position and decreases when farther from it. This behavior is due to the adaptive mechanism, which aims to maximize grasping capabilities only when the manipulator is close to the litter. When far from it, the system prioritizes minimizing rope tension. This adaptive behavior is achieved using a sigmoid function, with parameters $w = 1.0$, $\kgr = 1.0$, and $\kpos =1.0 $. 
The sigmoid function is detailed in the Methods section (see \sect{sec:AdaptiveCostWeight}), while the parameter selection is discussed in Supplementary Method~6.

The results, depicted in \fig{fig:ValExp_Traj}a, show that the robots properly follow their trajectories, while the resulting trajectory of the rope’s middle point (red line) remains close to the reference. In addition, \fig{fig:ValExp_Traj}b demonstrates that the aerial manipulator system successfully grasps the litter. Furthermore, \fig{fig:ValExp_cam}a presents the rope shape estimation pipeline at different time instants. The RGB image captured by the onboard camera (first column) is filtered to isolate the red color of the rope, resulting in a binary mask (second column). Using this mask, the rope’s point cloud is obtained. Finally, the parabolic shape is estimated from this point cloud and represented as a blue line. The estimated parameters are then used by the shape visual servoing controller. \fig{fig:ValExp_cam}b-bottom illustrates how the reference shape, computed by the rope-shape planner, is properly tracked. These estimated parameters are further utilized to determine the rope’s middle point and assess the precision of the complete pipeline. The temporal evolution of the rope’s middle point is depicted in \fig{fig:ValExp_cam}b-top, demonstrating that the control of the rope shape results in a middle point trajectory closely following the reference $\prefhook$.

\subsubsection{Ablation Study}

The success grasping presented before is in part achieved thanks to the optimal rope-shape planner, which aims to maximize the grasping capabilities of the rope close to the litter position. To show the importance of this weight of the planner during the operation, we perform a ablation study for different weights values $w$ and different trajectories. 

\begin{figure*}[t]
	\centering
	\includegraphics[width=2\columnwidth]{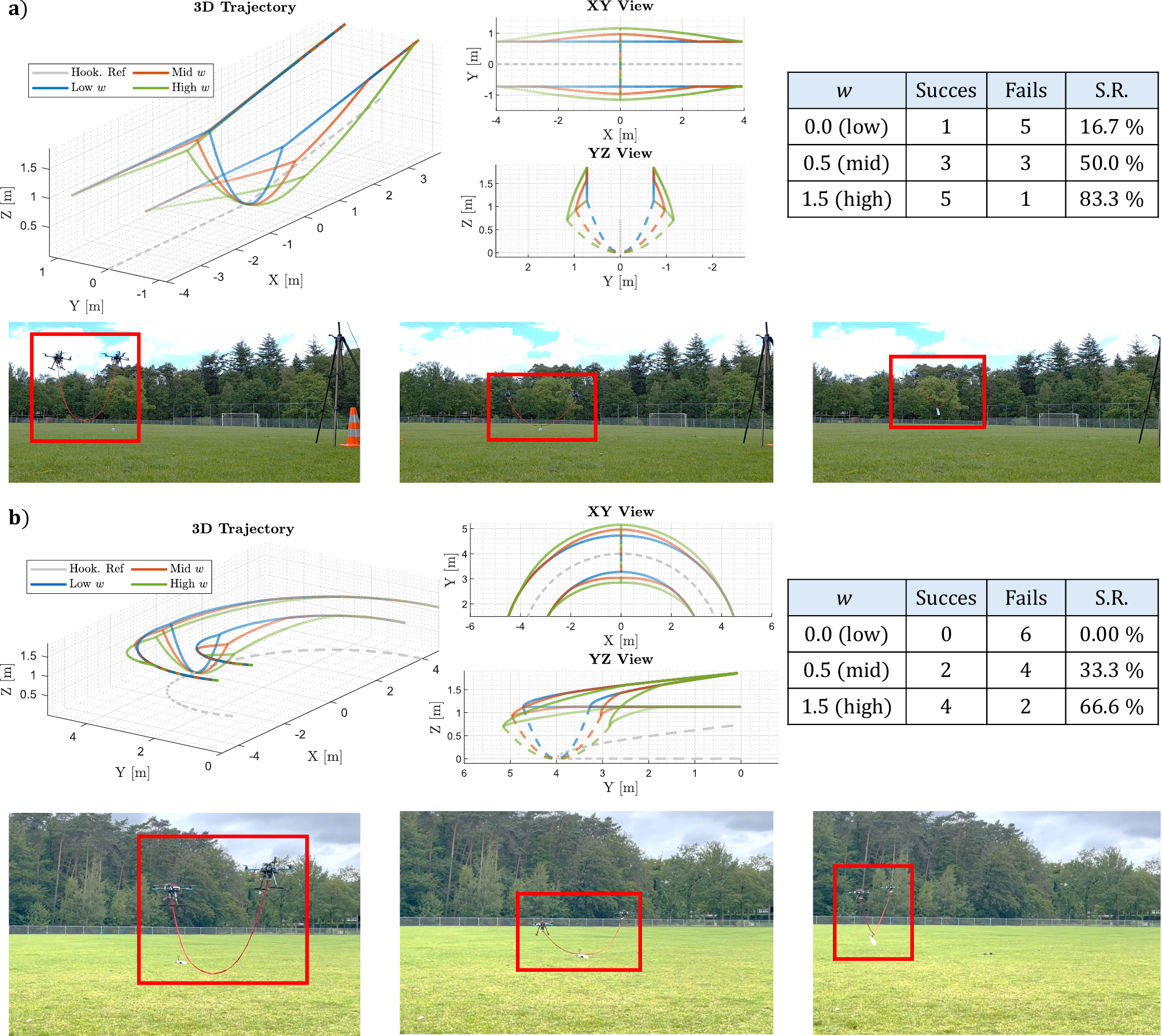}
    \caption{Ablation study in the grass field for different trajectories and weights $w$. a) Straight trajectory. b) Circular trajectory. Each case includes three snapshots and a summary table with the number of successes (bottle grasped), failures (bottle not grasped), and success rates.}\label{fig:Ablation}
\end{figure*}

The first trajectory is similar to the one presented before, a straight line in $\yW$, with an ascension phase after reaching $\plitter$. This reference trajectory $\prefhook$ is shown in \fig{fig:Ablation}a as a gray dashed line. The second trajectory is a semicircular trajectory, as represented in \fig{fig:Ablation}b. In this case, the litter position is $\plitter=[0.0,4.0,0.0]^\top\ \si{\meter}$, while the initial position of the rope middle point is $[4.0,0.0,0.0]^\top\ \si{\meter}$. In particular, the middle point follows a circular trajectory of center $[0.0,0.0,0.0]^\top\ \si{\meter}$, with a linear velocity of $0.50\ \si{\meter\per\second}$. After reaching the litter position $\plitter$, the middle point ascent with a vertical velocity of $0.10\ \si{\meter\per\second}$.

Both trajectories are evaluated for three different values of the optimization weight \(w\): a high value (\(w=1.5\)), a mid value (\(w=1.0\)), and a low value (\(w=0.0\)). In the case of the low value, the rope-shape planner does not consider the grasping function, so the rope tension becomes the only objective function in the optimization problem. As a result, the separation between the aerial robots remains constantly minimal throughout the entire trajectory. Specifically, the robots' trajectories have the shape shown in \fig{fig:Ablation} in blue lines.

The reference trajectories computed by the rope-shape planner are shown in \fig{fig:Ablation}. In particular, the high-value case (green lines) exhibits the greatest separation when the manipulator is close to \(\plitter\), while the low-value case (blue lines) maintains the minimum separation. To evaluate the grasping capabilities of each case, six flights are performed for each trajectory and for each weight. The results, in terms of success rate, are included in the tables of \fig{fig:Ablation}. These results show that higher weight values lead to improved grasping capabilities, as evidenced by the increased success rate. This occurs because a higher weight increases the importance of the grasping function in the optimization problem. This ablation study validates the significance of the weights in the rope-shape planner for enhancing the grasping capabilities of the aerial manipulator system.

\subsubsection{Floating litter collection in channel}

\begin{figure*}[t]
	\centering
	\includegraphics[width=2\columnwidth]{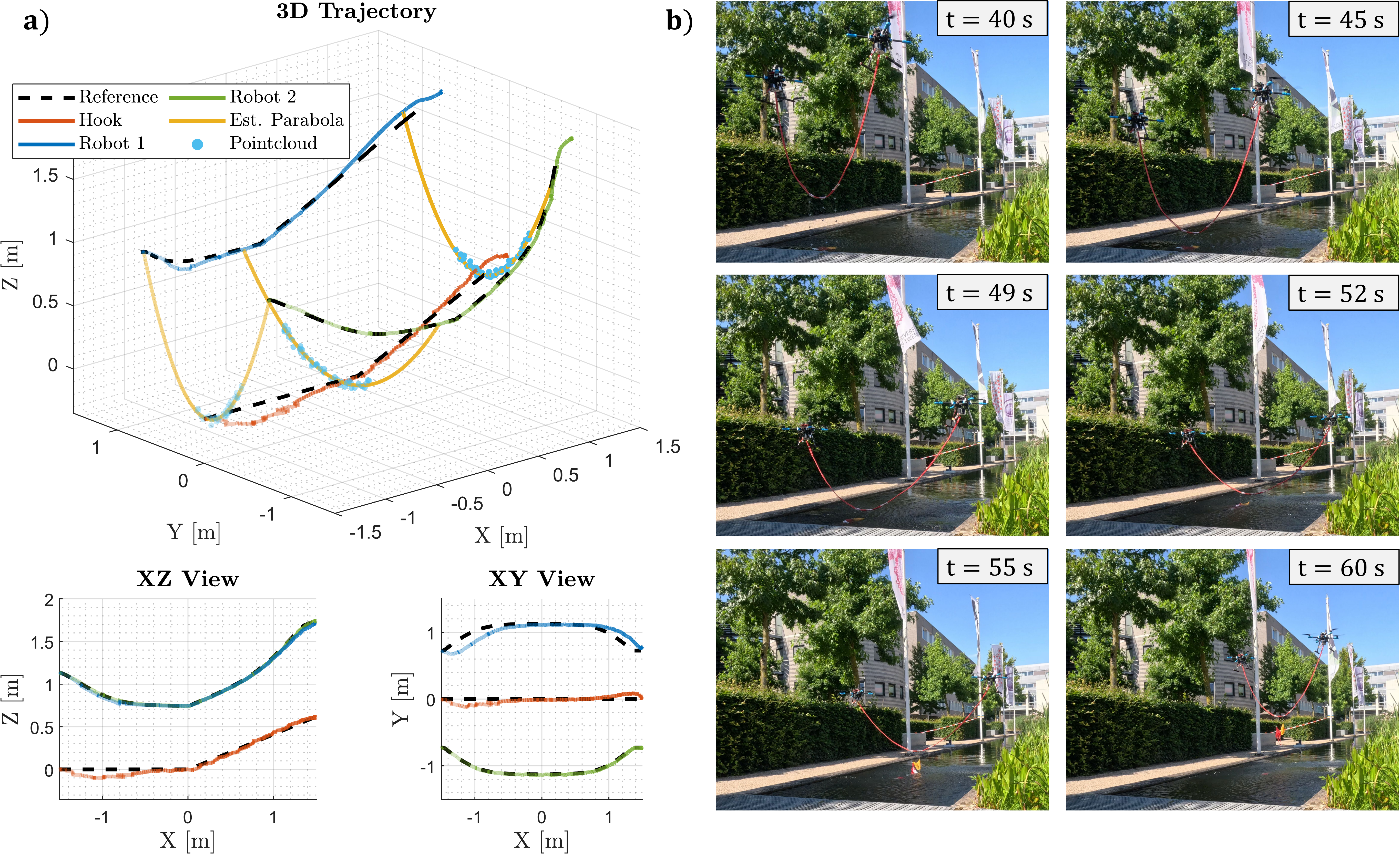}
    \caption{Results of the experiments in water channel. a) 3D and 2D views of the trajectories during the floating litter collection in channel. The transparency effect indicates the temporal progression of the trajectories, transitioning from transparent (initial) to solid (final) lines. The trajectories of robot 1 (blue line), robot 2 (green line) and hook (red line) are represented as solid lines, while the reference trajectories are represented as black dashed lines. In addition, the estimated parabola shape (yellow lines), and the pointcloud generated by the depth camera are represented for three different instants. b) Snapshots of the water channel experiment for six different instants. }\label{fig:water_traj}
\end{figure*}

The aerial manipulator system is finally validated through experiments in a water channel, where the hook-based tool is used to collect floating litters. In particular, we introduced three plastic bags in the water 
around a specific position $\plitter=[0.0,0.0,0.0]^\top \ \si{\meter}$ (see \fig{fig:task_def}). Due to the straight shape of the water channel, we define a straight reference trajectory $\prefhook$ to collect the floating litters, which is composed of a forward movement in the $\xW$ axis at $0.24\ \si{\meter\per\second}$, and an ascension movement in $\zW$ at $0.1\ \si{\meter\per\second}$ after reaching $\plitter$. With the reference trajectory $\prefhook$ and the litter position $\plitter$, the optimal planner computes the reference rope shape and the aerial robots trajectories. In this case, we use $w = 1.0$, 
$\kgr = 1.0$, and $\kpos = 1.0$ in the optimization problem. Although $w=1.5$ improved performance in the field experiments, in the water channel setup we use $w=1.0$ to limit robot separation and reduce the risk of lateral collisions with vegetation along the channel (see \fig{fig:water_traj}-b). The resulting reference trajectories are represented in \fig{fig:water_traj}-a. These experiments carried out with a $2\ \si{\meter\per\second}$ wind. The experimental setup in the water channel is detailed in the Methods section, while the video of the experiment is included in Supplementary Movie S2.

The floating litter collection operation is composed of different phases (see \fig{fig:water_traj}b). First, two pilots take off and manually fly the aerial robots until they are close to the initial position. After this, both aerial robots begin tracking the desired trajectories computed by the planner. During this phase, the aerial robots follow the planned trajectories, resulting in an open rope shape while the hook remains submerged to grasp the floating litter. Finally, after reaching \(\plitter\), the aerial robots ascend to retrieve the collected litter. All these phases of the operation are depicted in \fig{fig:water_traj}b, where the floating litter was successfully collected by the aerial robots.

\begin{figure*}[t]
	\centering
	\includegraphics[width=2\columnwidth]{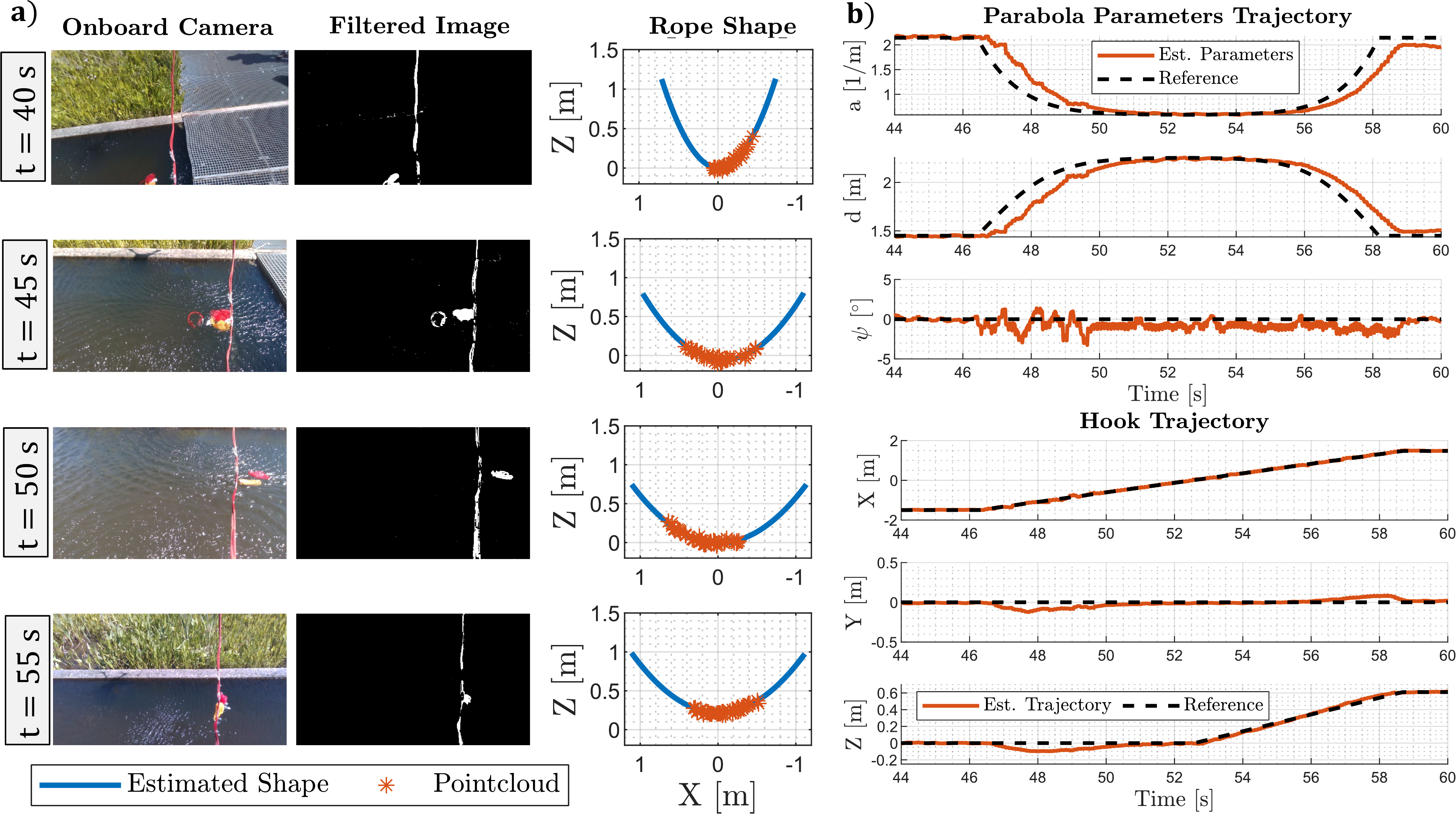}
    \caption{Parabola shape estimation during the  the experiments in water channel. a) Onboard camera view (first column), filtered image (second column) and filtered pointcloud and estimated parabola shape (third column). The three time instants corresponds with $t=20\ \si{\second}$ (first row), $t=25\ \si{\second}$ (second row), and $t=30\ \si{\second}$ (third row).  b) Rope trajectories during the validation experiment in the grass field. Bottom: reference hook trajectory (dashed black lines) and estimated trajectory (red lines). Top: parabola parameter reference (dashed black lines) and estimated parameters (red lines). The three shape parameters are the separation $\drefr$, the curvature $\arefr$ and the yaw angle $\psirefr$.}
    \label{fig:water_cam}
\end{figure*}

In addition, the rope shape is controlled throughout the entire operation. \fig{fig:water_cam}a illustrates the rope shape estimator pipeline at different time instants. Unlike the grass field experiments, in this case, some red floating litters are included in the binary mask during the filtering process. However, the point cloud is properly filtered, eliminating these points from the final point cloud. The parabola parameters estimated by this pipeline are shown in \fig{fig:water_cam}b-top, demonstrating that the shape is accurately controlled. Finally, using these parameters, the rope's middle point is estimated in \fig{fig:water_cam}b-bottom, which shows that the final hook trajectory closely follows the desired trajectory $\prefhook$. These results validate the proposed pipeline in real environments, where additional factors like water reflections may occur.

Furthermore, to demonstrate the influence of the weight of the optimal planner on the system's collecting capabilities, we compare two different scenarios in the water channel experiments. Specifically, the results for \(w=1.0\) are compared with the scenario in which grasping capabilities are not considered, which corresponds to selecting \(w=0\) in the planner. In this case, the reference trajectories of the robots remain parallel to the straight hook reference \(\prefhook\), resulting in minimal rope separation, as previously shown in the ablation study. We conducted six experiments for each scenario. The success rate for \(w=1.5\) was \(66.6\%\) (four experiments resulted in successful collection of at least one bag, while two did not), whereas for \(w=0.0\), the success rate was only \(16.7\%\) (one successful collection of at least one bag and five failures).

Compared to the grass field experiments, in the water setup, the use of hook tools allows for grasping the floating litter. In addition, since the litter is now floating on the water, we observed that downwash still slightly influences the position of the litter during the operation. In particular, in the low-weight scenario (\(w=0\)), the downwash significantly perturbs the position of the litter, as the robots are closer to the manipulation point. This causes the litter to drift on the water surface, and therefore, results in a low success rate. However, thanks to the increased separation when \(w=1.5\), the perturbation caused by downwash on the litter's position is greatly reduced, resulting in a higher success rate compared to the scenario with \(w=0\).

\section{Discussion}

Plastic pollution in oceans, rivers, and lakes has become a critical environmental issue, posing severe threats to marine ecosystems and biodiversity. Large quantities of plastic waste accumulate in water bodies, disrupting aquatic life and entering the food chain, potentially impacting human health. Addressing this challenge requires innovative and efficient solutions. In recent years, robotics-based approaches have been developed to enhance waste collection efforts in aquatic environments. Among these approaches, aerial robots have emerged as a promising technology for assisting in autonomous litter collection. Their ability to cover large areas rapidly, and their capacity to operate in complex environments make them valuable assets in water-cleaning missions. 

In this work, we present an integrated aerial robotic solution for floating litter collection from the water surface. Our solution consists of two aerial robots connected by a flexible rope with a hook-based tool installed at its midpoint, serving as a manipulator. The two aerial robots coordinate to control the shape of the rope manipulator by adjusting their trajectories accordingly. The use of two aerial robots helps reduce the downwash effect in the manipulation area, minimizing its impact on the position of the floating litter. Additionally, the rope shape is planned using a rope-shape planner that maximizes grasping capabilities while also minimizing tension in the rope. The desired rope trajectory is then controlled by the aerial robots through a visual servoing strategy, which approximates the rope shape as a parabola. The complete system has been validated through various grasping tasks, where the influence of the planner's weight parameters was also evaluated. Furthermore, several water channel experiments demonstrate that our approach provides an effective solution for floating litter collection.

The current solution assumes that the litter position is provided and remains fixed throughout the operation. As future work, we aim to integrate our aerial system with advanced floating litter detection techniques, enabling fully autonomous operation. In this case, the system would be capable of detecting floating litter and planning a trajectory to grasp it. Additionally, this approach would allow our solution to be applied in moving water environments, such as rivers, rather than being limited to still waterways like channels or lakes. The system could dynamically replan its trajectory to account for the movement of the litter, increasing its adaptability and effectiveness. Another future direction is to test the system with alternative lightweight end-effectors, such as nets instead of hooks, to enable the grasping of different types of objects, including cans or bottles.

\section{Methods}
\subsection{Aerial system manufacturing}

\begin{figure*}[t]
	\centering
	\includegraphics[width=2\columnwidth]{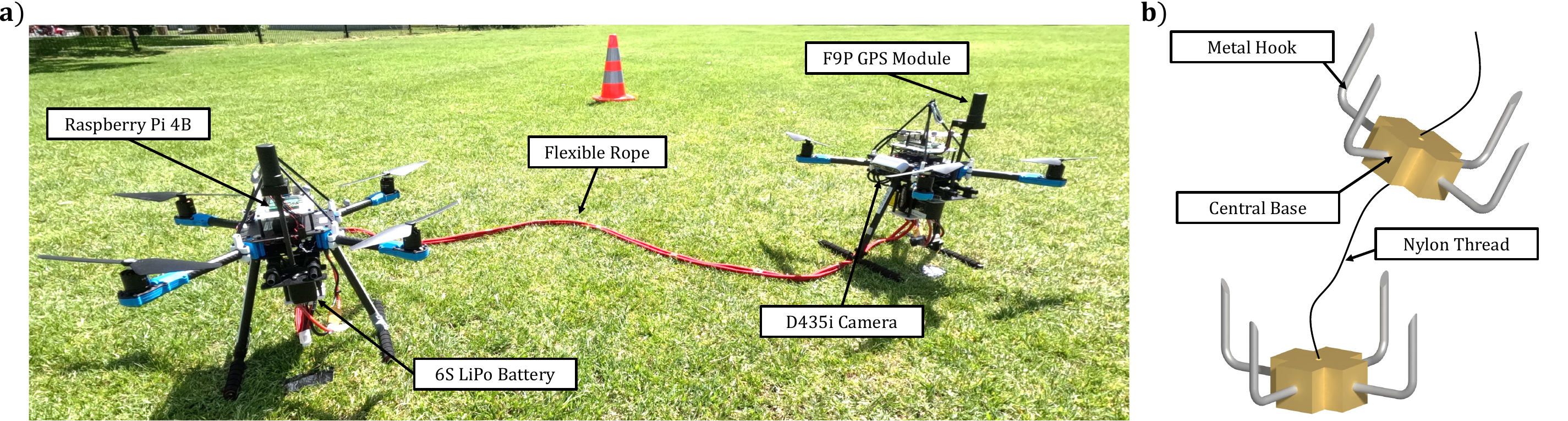}
    \caption{Aerial system for floating litter collection. a) Main components of the aerial system, including two commercial Holybro X500 quadrotors and a flexible red rope. b) Components of the hook-based tool for floating litter collection.}\label{fig:aerial_system}
\end{figure*}

The aerial system consists of two commercial Holybro X500 quadrotors and an Orion 500 red rope, as shown in \fig{fig:aerial_system}.  

Each quadrotor is equipped with a Pixhawk 6C flight controller running the PX4 autopilot as the inner position controller. This autopilot implements a PID-cascade structure for both position and attitude controllers (see Supplementary Method 1 and Supplementary Figure 1 for more information). A reference frame $\FRi=\{\oRi, \xRi, \yRi, \zRi\}$ with $i=1,2$, where $\oRi$ is the origin and $\xRi, \yRi, \zRi$ three orthonormal axis, is fixed to the Center of Mass (CoM) of the $i^{th}$ robot. Similarly, define an inertial world-fixed frame $\FW=\{\oW, \xW, \yW, \zW\}$ located at the RTK-GPS base, with $\xW$ aligned with the Earth's north and $\zW$ opposite to the direction of gravity. The current position $\posRA\in\R{3}$ and $\posRB\in\R{3}$ of each robot in the inertial reference frame is obtained from an F9P GPS module, while the attitude of $\FRA$ and $\FRB$ w.r.t. $\FW$, represented by the rotation matrices $\wRrA\in\SO{3}$ and $\wRrB\in\SO{3}$ respectively, is estimated fusing the onboard IMU and the GPS magnetometer. Additionally, the aerial robot~1 is equipped with an Intel D415i camera to provide visual and depth data for rope shape estimation. The RGB-D camera is mounted on the platform using a 3D-printed piece. Using this piece, the camera is positioned at a downward-facing angle of $45\unit{\degree}$ with respect to the horizontal plane of the robot frame. Each aerial platform is powered by a 4500 mAh 6S LiPo battery. A detailed description of the hardware architecture of the aerial robot is available in Supplementary Method 2 and Supplementary Figure 2.  

Furthermore, the quadrotors include a Raspberry Pi 4B as an onboard computer, where the entire control framework runs, except for the inner controllers of the aerial robots. The control framework is implemented using ROS, which facilitates communication between the different components of the framework and the needed hardware.The visual servoing controller, the parabola parameter estimation node, and the point cloud processing node run onboard Robot 1, while the inner controllers and trajectory execution run on each robot’s PX4 autopilot. The planned trajectories are published from a ground computer. A ROS multimaster architecture is used to enable communication between nodes running on the ground computer and the onboard computers of both robots, while maintaining three independent ROS masters for increased robustness.  Supplementary Method 3 and Supplementary Figure 3 detail the software architecture.

The selected flexible rope is made of polyester, with a total length of $5.6\ \si{\meter}$ and a diameter of $8\ \si{\milli\meter}$. To increase the effective width of the rope and enhance its visibility for the depth sensor, the rope is folded in half. As a result, the rope manipulator has a final length of $\Lrope = 2.8\ \si{\meter}$ and a width of $16\ \si{\milli\meter}$. The rope extremes are connected to the robot frames with plastic bridges, guaranteeing a spherical passive joint between the robots and the rope. The attachment points are at a distance $\zatt=0.12\ \si{\meter}$ along $\zRi$ from the robots CoM. 

Additionally, a hook-based tool, composed of nylon threads with hooks (see \fig{fig:aerial_system}), is installed at the midpoint of the rope. The hooks are fabricated by inserting an iron wire through two lateral holes in a 3D-printed central base. The wires are then bent to form the shape of a hook. A nylon thread is used to attach all the individual hooks. In our aerial system, each nylon thread holds five hooks spaced $0.10\ \si{\meter}$ apart, with three nylon threads arranged $0.15\ \si{\meter}$ apart from each other in the center of the rope (see \fig{fig:task_def}). As a result, the length of rope segment with hooks installed is $\Lhook=30\ \si{\centi\meter}$. With this setup, the total mass of the flexible manipulator is $\mrope=0.25\ \si{\kilo\gram}$, including the weight of the hooks-based tool.


\subsection{Rope-shape planner}\label{sec:rop-planner}

\begin{figure*}[t]
	\centering
	\includegraphics[width=2\columnwidth]{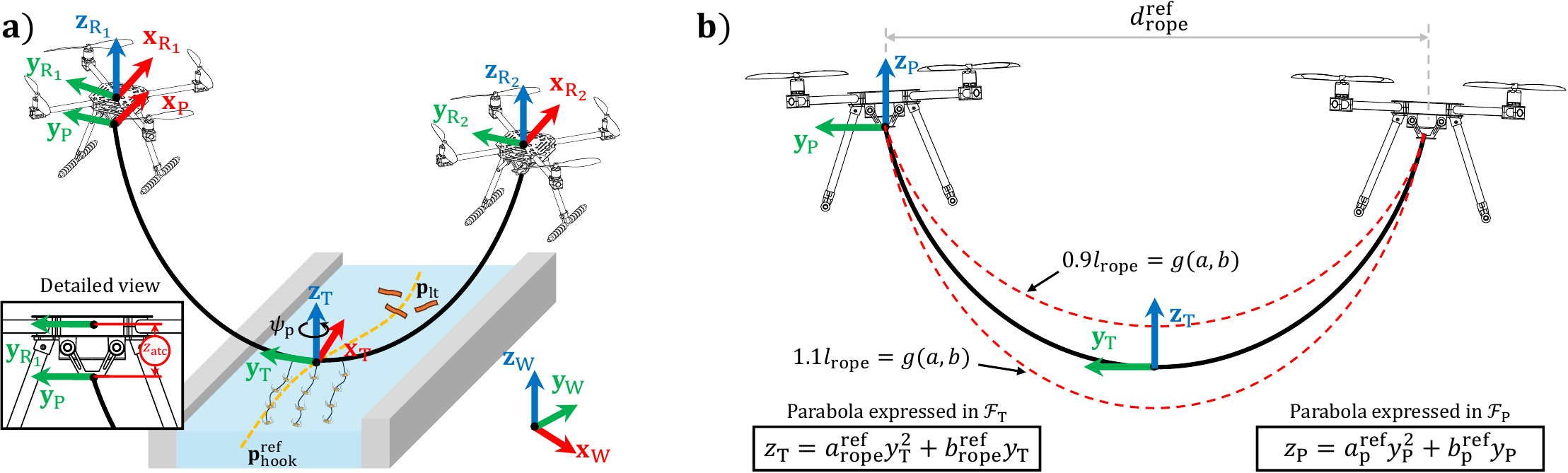}
    \caption{Frames used in this work. a) Main frames and position vectors. b) Parabola equations for the $\FT$ frame, attached on the rope middle point, and for the $\FVS$ frame, located at the attachment point of robot 1. The two constraint shapes of the optimization problem \eqn{eq:cont_l} are represented by red dashed lines.}\label{fig:frames}
\end{figure*}

 We introduce a rope tool frame $\FT$, defined by its origin $\oT$ and axes $\xT, \yT,$ and $\zT$. This frame is positioned at the manipulation point of the rope, where the hook-based tool is installed, with $\xT$ directed normal to the containing parabola plane and $\yT$ parallel to the ground. \fig{fig:frames}a represents a detailed representation of all reference frames used in this work, including the two body frames, $\FRA$ and $\FRB$, attached to the robots.

The reference trajectory $\prefhook = [\xrefhook,\yrefhook,\zrefhook] \in \R{3}$, expressed in $\FW$, describes the trajectory of the tool point of the parabola, i.e., $\oT$. The rope-shape planner computes the desired shape of the rope for a predefined reference trajectory. The rope shape is parameterized by the parabola $z(y,t) = a(t)y^2 + b(t)y$, whose curve $z(y,t)$ is defined in the parabola frame $\FT$ (see \fig{fig:frames}b). This parabola is characterized by four parameters: $a \in \R{}$, $b \in \R{}$, $\phi\in\R{}$ and $\psi \in \R{}$. The parameters $a$ and $b$ determine the shape of the parabola in the $\yT\zT$ plane, while $\phi$ and $\psi$ represent the roll and yaw angles of its containing plane, respectively. The parameters $a$, $b$ and $\psi$ can be controlled using the visual servoing controller (see \sect{sec:vis_ser}). However, the roll angle $\phi$ can not be controlled through the trajectories of the robots. Nevertheless, since the cable does not directly interact with the water and the movements are slow, we assume this angle remains close to zero. The rope-shape planner computes the desired value for the other three parameters ($\arefr$, $\brefr$ and $\psirefr$) for a given reference trajectory $\prefhook \in \R{3}$.  

Since the manipulation area coincides with the midpoint of the rope, we enforce a symmetric shape with respect to $\yP$ to ensure that this area remains the closest point to the water surface. This condition is satisfied when $\brefr(t) = 0$.


To maximize hook coverage, we ensure that the $\yT\zT$ plane, assumed to be vertical since the roll angle is considered negligible, is perpendicular to the horizontal projection of $\prefhook$. This is achieved mathematically by:
\begin{equation}
    \label{eq:psiRope}
    \psirefr(t)=\arctan\left(\dfrac{\ydrefh(t)}{\xdrefh(t)}\right),
\end{equation}
which guarantees that the parabola plane remains perpendicular to the horizontal trajectory defined by $\xrefhook(t)$ and $\yrefhook(t)$.

The remaining parameter $\arefr(t)$, which defines the curvature of the parabola, is obtained by solving an optimization problem. To do this, we introduce the separation $\drefr(t)\in\R{+}$ between the robots as a new optimization variable (see \fig{fig:frames}b). With all this, the optimization problem is:
\begin{mini!}|s|
	{\arefr(t),\drefr(t)}{J(\arefr(t),\drefr(t),\prefhook(t)-\plitter(t))}
	{\label{eq:opt}}{}
	\addConstraint{\Lrope=\gfun(\arefr(t),\drefr(t))} \label{eq:lengthFun}
    \addConstraint{\dmin \leq \drefr(t) \leq \dmax} \label{eq:limd}
	\addConstraint{\arefr(t) \geq 0} \label{eq:limita},
\end{mini!}
where $\arefr$ and $\drefr$ are the optimization variables, and $\plitter\in\R{3}$ is the litters position, which is supposed known. As the reference trajectory $\prefhook$ is previously defined, the optimization planner can be solved offline. In our case, we solve this optimization problem using $Matlab$.

\subsubsection{Constraints}

The equality constraint \eqref{eq:lengthFun} imposes the relation between the separation $\drefr$ and the curvature $\arefr$ through the total length of the rope $\Lrope$, with $\gfun:\R{}\rightarrow\R{+}$. This function is given by:
\begin{align}
&\Lrope=\gfun(\arefr,\drefr)=\arefr\drefr\sqrt{1+(\arefr\drefr)^2}+\nonumber \\ &+\arcsinh (\arefr\drefr), \label{eq:gfun}
\end{align}
whose deduction is detailed in Supplementary Method 4.

Moreover, the inequality constraint \eqref{eq:limd} limits the separation between the robots. The minimum separation $\dmin\in\R{+}$ establishes a safety distance between robots, and it is specified by the user. Moreover, the maximum separation $\dmax\in\R{+}$ is given by:
\begin{equation}
    \dmax = \min(\dmaxh,\dmaxf),
\end{equation}
where $\dmax$ selects between the separation $\dmaxh\in\R{+}$, which results in the limit flight height $\hmin\in\R{+}$, and the separation $\dmaxf\in\R{+}$, which results in the maximum roll angle $\phimax\in\R{+}$ for compensating rope tension. In particular, $\hmin$ represents the minimum flight height (security distance to the floor) allowed for the robots, while $\phimax$ represents the maximum roll angle allowed for the robots. Both parameters are defined by the operator based on the desired safety and control constraints. The corresponding distances, $\dmaxh$ and $\dmaxf$, are computed prior to the optimization problem, based on the selected values of $\hmin$ and $\phimax$. Supplementary Method 4 details the computation of $\dmaxh$ and $\dmaxf$.

Finally, the inequality constraint \eqn{eq:limita} ensures solutions with positive curvature, thereby guaranteeing a reference shape where the rope remains below the robots. Note that if $\arefr<0$, the resulting parabolic shape would correspond to a scenario where the rope is positioned above the robots. A detailed mathematical derivation of all the constraints is provided in Supplementary Method 4.

\subsubsection{Objective function}
The planner aims to minimize the objective function $J$:
\begin{equation}
    J=\ftension(\arefr,\drefr)+\wgrasp(\prefhook(t)-\plitter(t))\fgrasp(\arefr,\drefr),
\end{equation}
which is composed of two main terms: (i) minimize the tension in the rope extremes (term $\ftension\in\R{+}$); and (ii) minimize the unused hook width (term $\fgrasp\in\R{+}$). 

The tension $\ftension$ of the rope in its extremes is given by:
\begin{equation}
    \ftension=\tension=\dfrac{\mrope g\sqrt{1+(\arefr\drefr)^2}}{2\arefr \drefr},
\end{equation}
where $\g$ is the gravity, and the mass of the collected litters is neglected compared with the mass of the rope. 

The unused hook width $\fgrasp$ is determined by:
\begin{equation}
    \fgrasp=\Lhook-\dhook(\Lhook,\arefr),
\end{equation}
where the horizontal width $\dhook$ covered by the hooks is obtained by solving the equation $\Lhook=g(\arefr,\dhook)$. Supplementary Method 5 details the derivation of both expressions.

\subsubsection{Adaptive cost weight}
\label{sec:AdaptiveCostWeight}

The objective function $\fgrasp$ is weighted by ${\wgrasp(\prefhook(t)-\plitter(t)):\R{3}\rightarrow\R{+}}$, which varies depending on the distance to the litter. This adaptation allows to only consider this objective function when the hooks are closed to the target litter, located at $\plitter$. A sigmoid is used to modulate the weight as:
\begin{equation}
    \wgrasp=\dfrac{w}{1+\exp{(-\kgr\left(\kpos-\NormTwo{\prefhook-\plitter}\right))}},
\end{equation}
where $w\in\R{+}$ is a constant weighting factor, $\kgr\in\R{+}$ modifies the shape of the sigmoid and $\kpos\in\R{+}$ represents the position where the sigmoid pass from 0 to $w$. The impact of $\wgrasp$ on the optimal planner has been studied through simulations and is available in Supplementary Method 6 and Supplementary Figure 4.

\subsection{Aerial robots planned trajectories} With $\arefr(t)$ and $\drefr(t)$ from the optimal problem \eqn{eq:opt}, $\brefr(t)=0$ from the symmetric parabola shape condition, and $\psirefr(t)$ from \eqn{eq:psiRope}, we compute the planned trajectories for both robots as:
\begin{subequations} \label{eq:traj_uavs}
	\begin{align}
		\pplRA & =\zatt\zW+ \prefhook + \Rz(\psirefr)(\arefr\left(\frac{{\drefr}}{2}\right)^2\zT+\nonumber \\&+\frac{\drefr}{2}\yT), \\
        \psiplRA& = \psirefr, \\
        \pplRB & = \zatt\zW+\prefhook +  
        \Rz(\psirefr)(\arefr\left(\frac{{\drefr}}{2}\right)^2\zT+
        \nonumber\\ &-\frac{\drefr}{2}\yT), \\
        \psiplRB& = \psirefr,
	\end{align}
\end{subequations}
where $\pplRA, \pplRB \in \R{3}$ denote the planned trajectories of robot~1 and robot~2, respectively. Similarly, $\psiplRA, \psiplRB \in \R{}$ represent their planned yaw angles. Additionally, $\Rz(\psirefr) \in \SO{3}$ denotes the rotation matrix around $\zW$, which transforms coordinates from $\FT$ to $\FW$. Note that the term $\zatt\zW$ compensates the distance between robots CoM and the attachment points, assuming small roll and pitch angles of the robots. The planned yaw trajectories in \eqn{eq:traj_uavs} guarantees that the rope remains in the lateral of the robots, enhancing to a properly data acquisition with the RGB-D camera installed on the right side of robot 1. 



\subsection{Rope Shape Visual Servoing}\label{sec:vis_ser}
The rope shape visual servoing controller is developed based on \cite{smolentsev2024shape}, including some modifications. In this controller, the estimated parabola parameters $\spab=[\avs,\bvs,\psivs]^\top\in\R{3}$ (see \sect{sec:sh_estimation} for the rope shape estimation) are described in a new frame $\FVS$. This new frame $\FVS$ has the same orientation as $\FT$, but with its origin $\oVS$ located at the attachment point of robot 1 (see \fig{fig:frames}a). The parabola parameter error $\es\in\R{3}$ is computed as $\es=\spab-\sref$, where $\sref=[\avsref,\bvsref,\psivsref]^\top\in\R{3}$ must be expressed in the new frame $\FVS$. Therefore, the planned shape provided by the planner must be converted into the frame $\FVS$. Considering $\brefr(t)=0$, the parabola reference for the visual servoing controller is:
\begin{subequations} \label{eq:pb-param}
    \begin{align}
        \avsref&=\arefr,\\
        \bvsref&=-\arefr\drefr,\\
        \psivsref&=\psirefr.
    \end{align}
\end{subequations}

With this, the visual servoing controller is:
\begin{equation}\label{eq:vs_pi}
    \vrel=\kc \Mmat(\psivs) \es + \ki \Mmat(\psivs) \sum_{i=0}^{n_w}\esi,
\end{equation}
where $\vrel\in\R{3}$ is the relative velocity between robot 2 and robot 1, and $\Mmat\in\R{3\times3}$ is the interaction matrix, which is computed as detailed in \cite{smolentsev2023shape}. Finally, $\kc\in\R{+}$ and $\ki\in\R{+}$ are the proportional and integral gains of the PI controller. In our case, the integral term only considers the past errors $\esi$ in a sliding window of size $n_w\in\N$.

We use \eqref{eq:vs_pi} to compute the correction terms $\DeltaprefRA\in\R{3}$ and $\DeltaprefRB\in\R{3}$  as: 
\begin{align}
&\DeltaprefRA=\Deltapvs=\int\dfrac{\vrel}{2}dt \ \ \text{and} \ \ \DeltaprefRB=\nonumber\\ &=-\Deltapvs=-\int\dfrac{\vrel}{2}dt,
\end{align}
where $\vrel$ is computed using \eqn{eq:pb-param}. $\DeltaprefRA\in\R{3}$ and $\DeltaprefRB\in\R{3}$ are then given to the robots' controller as shown in \fig{fig:control_sh}. Compared with the control law proposed in \cite{smolentsev2023shape,smolentsev2024shape}, our controller distributes equally the needed relative velocity between both robots. Thanks to this, the reference trajectory of the hooks remains unchanged, as the trajectory of the rope's middle point stays the same due to the vertical symmetry of the control action.



\subsection{Parabola shape estimation}\label{sec:sh_estimation}


The proposed pipeline for the parabola shape estimation is based on \cite{smolentsev2024shape}, incorporating some improvements. It is composed of two main stages: (i) the pointcloud generation; and (ii) the parameters estimation.

\subsubsection{Pointcloud generation} The pipeline starts with the data acquisition from the RGB-D camera. The rope is then identified in the RGB image by segmenting the cable based on its color, red in our case. A binarization process is performed to further isolate the rope in the RGB image. The resulting binary mask is applied to the aligned depth data, retaining only the rope’s depths. With this depth data, we construct a point cloud, expressed in the camera frame, using the intrisic camera parameters. Then the pointcloud is downsampled using a voxelized grid approach. The centroid of each voxel is then computed, and only its coordinates are used further. This downsampled pointcloud, composed of $\np\in\N$ points and expressed in the camera frame $\FC$, is subsequently transformed into the world frame $\FW$. To do this, we use the current position of robot 1, $\posRA$, and the static transformation between the robot 1 frame $\FRA$ and the camera frame $\FC$. This static transformation is obtained using the hand-eye calibration method. More details about these steps are available in \cite{smolentsev2024shape}.

At this stage, we refine the algorithm from \cite{smolentsev2024shape} to enhance the robustness of the estimation. The current positions of both attachment points $\pattA,\pattB\in\R{3}$ are also incorporated into the rope’s pointcloud, as they represent its endpoints. The resulting point cloud $\PW$ is represented by the set:
\begin{equation}
    \PW=\{\pattA,\pIn,\pattB\},
\end{equation}
where the attachment positions are computed using $\zatt$ as $\pattA=\posRA-\zatt\zB$ and $\pattB=\posRB-\zatt\zB$; furthermore, $\pIn\in\R{3},\ \forall i \in[1,\np]$ represents the position of the pointcloud points generated by the depth camera.

\subsubsection{Parameters estimation}
With this pointcloud $\PW$, the normal direction $\nvec=[n_x,n_y,n_z]^\top\in\R{3}$ and the parameter $d\in\R{}$ of the best-fitting plane are estimated. These parameters, $\nvec$ and $d$, define the plane equation $\nvec^\top \mathbf{p} + d = 0$, where $\mathbf{p}\in\R{3}$ is a point lying on the plane. Therefore, the parameters of the best-fitting plane are computed by solving:
\begin{mini!}|s|
	{\nvec,d}{(\nvecT \pattA +d)^2+ (\nvecT \pattB+d)^2 + \sum_{i=1}^{\np} (\nvecT \pIn+d)^2,}{}{}\label{eq:nvec_opti}
\end{mini!}
using the RANSAC algorithm. With $\nvec$ from \eqn{eq:nvec_opti}, the roll $\phirope$ and yaw $\psirope$ angle of the parabola plane are computed as: 
\begin{equation}
    \psivs=\arctan\left(\dfrac{n_x}{n_y}\right)\ \ \text{and}\ \ \phivs = \arctan\left(\dfrac{n_z}{\sqrt{n_x^2+n_y^2}}\right).
\end{equation}

Then, to estimate the parameters $\avs,\bvs,\cvs$ of a general parabola $y=\avs z^2+\bvs z+\cvs$, the point cloud $\PW$ is transformed into the frame $\FVS$, resulting in the point cloud $\PT$:
\begin{equation}
    \PT=\{\zeroThree,\pUnoT,\dots,\pInT,\dots,\pNT,\pattBT\},
\end{equation}
where $\pattBT=[\xAtB,\yAtB,\zAtB]^\top\in\R{3}$ is the position of the attachment point of the robot 2, expressed in the frame $\FVS$. In addition, the first element is $\zeroThree=[0,0,0]^\top\ \si{\meter}$, as the attachment point $\pattA$ coincides with the origin $\oVS$. 

 With this, we solve the following optimization problem to estimate the parabola parameters:
\begin{align}
	&\min_{\avs,\bvs,\cvs}
    \begin{matrix}&\wmeas \cvs + \sum_{i=1}^{\np} \wdepth(\avs y_i^2+\bvs y_i
    +\cvs-z_i)+\\&+ \wgps(\avs\yAtB^2+\bvs\yAtB+\cvs -\zAtB)\\\end{matrix} \\ & \text{s.t. \ \ }{0.9\;\Lrope\leq g(\avs,\bvs,\yAtB)\leq1.1\;\Lrope}\label{eq:cont_l}
\end{align}
 where $\wgps,\wdepth,\wmeas\in\R{+}$ are the weights of the different terms, and are selected based on the precision of each measurement source (higher precision higher weight). The first one, $\wmeas$, is related with the precision of the manual measurement of the attachment point on robot 1. The second one, $\wdepth$, is related with the precision of the depth sensor. Finally, $\wgps$ is related with the precision of the onboard GPS module of robot 2, as it is used in the computation of $\pattB$. 
 Supplementary method~7 discusses the selection of this optimization weights. Finally, the inequality constraints \eqn{eq:cont_l} impose solutions where the total length of the curve is close to the real rope length $\Lrope$. The optimal problem is solved using CasADi \cite{Andersson2019}.

 Compared with \cite{smolentsev2024shape}, where the point $\pattA$ is directly imposed into the parabola equation by $\cvs=0$, our approach accounts for sensor uncertainty in the optimization problem through weights. In addition, thanks to the use of the length constraint \eqn{eq:cont_l}, we ensure a solution close to the real one, even without depth sensor data. These features highlight the improved robustness of our proposed approach.
 
Finally, a standard Kalman filter is applied to the estimated parabola parameters $\spab=[\avs,\bvs,\psivs]^\top$ to reduce noise.


\subsection{Experimental setup}

Both experimental set up, grass field and water channel, have the same components. The ground segment is composed of a ground computer connected to the RTK F9P base. The RTK corrections are sent to the robots through the QGround control station. The robots and the ground computer are connected to the same local network, created by a router. The ground computer commands the planned trajectories of the robots. 
Both robots are controlled by pilots only during take-off and landing maneuver. The pilots must have the RC on hands during the whole experiments, in case of problems. Supplementary Figure 5 and Supplementary Figure 6 present the main components of both setup.

\bibliographystyle{IEEEtran}
\bibliography{biblio}

\begin{thebibliography}{10}
\providecommand{\url}[1]{#1}
\csname url@samestyle\endcsname
\providecommand{\newblock}{\relax}
\providecommand{\bibinfo}[2]{#2}
\providecommand{\BIBentrySTDinterwordspacing}{\spaceskip=0pt\relax}
\providecommand{\BIBentryALTinterwordstretchfactor}{4}
\providecommand{\BIBentryALTinterwordspacing}{\spaceskip=\fontdimen2\font plus
\BIBentryALTinterwordstretchfactor\fontdimen3\font minus \fontdimen4\font\relax}
\providecommand{\BIBforeignlanguage}[2]{{%
\expandafter\ifx\csname l@#1\endcsname\relax
\typeout{** WARNING: IEEEtran.bst: No hyphenation pattern has been}%
\typeout{** loaded for the language `#1'. Using the pattern for}%
\typeout{** the default language instead.}%
\else
\language=\csname l@#1\endcsname
\fi
#2}}
\providecommand{\BIBdecl}{\relax}
\BIBdecl

\bibitem{horejs2020solutions}
C.~Horejs, ``Solutions to plastic pollution,'' \emph{Nature Reviews Materials}, vol.~5, no.~9, pp. 641--641, 2020.

\bibitem{sonke2024global}
J.~Sonke, A.~Koenig, T.~Segur, and N.~Yakovenko, ``Global environmental plastics dispersal under oecd policy scenarios towards 2060,'' 2024.

\bibitem{helinski2021ridding}
O.~K. Helinski, C.~J. Poor, and J.~M. Wolfand, ``Ridding our rivers of plastic: A framework for plastic pollution capture device selection,'' \emph{Marine pollution bulletin}, vol. 165, p. 112095, 2021.

\bibitem{koelmans2022risk}
A.~A. Koelmans, P.~E. Redondo-Hasselerharm, N.~H.~M. Nor, V.~N. de~Ruijter, S.~M. Mintenig, and M.~Kooi, ``Risk assessment of microplastic particles,'' \emph{Nature Reviews Materials}, vol.~7, no.~2, pp. 138--152, 2022.

\bibitem{kozlov2025your}
M.~Kozlov, ``Your brain is full of microplastics: are they harming you?'' \emph{Nature}, vol. 638, no. 8050, pp. 311--313, 2025.

\bibitem{honingh2020urban}
D.~Honingh, T.~Van~Emmerik, W.~Uijttewaal, H.~Kardhana, O.~Hoes, and N.~Van~de Giesen, ``Urban river water level increase through plastic waste accumulation at a rack structure,'' \emph{Frontiers in earth science}, vol.~8, p.~28, 2020.

\bibitem{rivercleaning}
``River cleaning plastic,'' \url{https://rivercleaning.com/}.

\bibitem{seads}
``Blue barriers by seads: Sea defence solutions,'' \url{https://www.seadefencesolutions.com/}.

\bibitem{watergoat}
``Watergoat,'' \url{https://www.watergoat.org/}.

\bibitem{oceancleanup}
``Interceptor barrier by the oceancleanup,'' \url{https://theoceancleanup.com/rivers/}.

\bibitem{noria}
``Circleaner by noria sustainable innovators,'' \url{https://www.noria.earth/}.

\bibitem{lee2024optimal}
J.~Lee, S.~Roh, J.~Im, M.~Kim, T.~Kim, and S.~Yoo, ``Optimal design of a floating waste-collecting robot utilizing vortex phenomena,'' \emph{IEEE Access}, 2024.

\bibitem{seabin}
``Seabin by the seabin foundation,'' \url{https://seabinfoundation.org/}.

\bibitem{bubblebarrier}
``The great bubble barrier,'' \url{https://thegreatbubblebarrier.com/}.

\bibitem{10165589}
C.~Ruixi, C.~Haoming, L.~Kexin, L.~Xintong, X.~Qiao, and Y.~Jinhan, ``Research and development of recycling boats for floating garbage,'' in \emph{2023 IEEE 3rd International Conference on Information Technology, Big Data and Artificial Intelligence (ICIBA)}, vol.~3, 2023, pp. 1229--1235.

\bibitem{zhu2022smurf}
J.~Zhu, Y.~Yang, and Y.~Cheng, ``Smurf: A fully autonomous water surface cleaning robot with a novel coverage path planning method,'' \emph{Journal of Marine Science and Engineering}, vol.~10, no.~11, p. 1620, 2022.

\bibitem{seavax}
``Seavax,'' \url{https://www.oceansplasticleanup.com/SeaVax_RiverVax/SeaVax_RiverVax_Projects_Overview.htm}.

\bibitem{wasteshark}
``the wasteshark by ranmarine,'' \url{https://www.ranmarine.io/products/wasteshark/}.

\bibitem{jellyfishbot}
``Jellyfishbot by iadys,'' \url{https://www.jellyfishbot.io/maritime-river-lake-water-sites/}.

\bibitem{zhang2023spiral}
Y.~Zhang, Z.~Huang, C.~Chen, X.~Wu, S.~Xie, H.~Zhou, Y.~Gou, L.~Gu, and M.~Ma, ``A spiral-propulsion amphibious intelligent robot for land garbage cleaning and sea garbage cleaning,'' \emph{Journal of Marine Science and Engineering}, vol.~11, no.~8, p. 1482, 2023.

\bibitem{muthusamy2024novel}
S.~Muthusamy, S.~Duraisamy, M.~Ramachandran, J.~Karthikeyan, J.~I. David, H.~K. Settu, and A.~Rathinasamy, ``A novel method for design and development of hybrid land and water buoyancy trash collecting robot,'' in \emph{2024 International Conference on Intelligent and Innovative Technologies in Computing, Electrical and Electronics (IITCEE)}.\hskip 1em plus 0.5em minus 0.4em\relax IEEE, 2024, pp. 1--5.

\bibitem{geraeds2019riverine}
M.~Geraeds, T.~van Emmerik, R.~de~Vries, and M.~S. bin Ab~Razak, ``Riverine plastic litter monitoring using unmanned aerial vehicles (uavs),'' \emph{Remote Sensing}, vol.~11, no.~17, p. 2045, 2019.

\bibitem{zoric2024towards}
F.~Zoric, A.~Franchi, M.~Orsag, Z.~Kovacic, and C.~Gabellieri, ``Towards instance segmentation-based litter collection with multi-rotor aerial vehicle,'' in \emph{2024 International Conference on Unmanned Aircraft Systems (ICUAS)}.\hskip 1em plus 0.5em minus 0.4em\relax IEEE, 2024, pp. 631--637.

\bibitem{gabellieri2023differential}
C.~Gabellieri and A.~Franchi, ``Differential flatness and manipulation of elasto-flexible cables carried by aerial robots in a possibly viscous environment,'' in \emph{2023 International Conference on Unmanned Aircraft Systems (ICUAS)}.\hskip 1em plus 0.5em minus 0.4em\relax IEEE, 2023, pp. 963--968.

\bibitem{abhishek2021towards}
V.~Abhishek, V.~Srivastava, and R.~Mukherjee, ``Towards a heterogeneous cable-connected team of uavs for aerial manipulation,'' in \emph{2021 American Control Conference}.\hskip 1em plus 0.5em minus 0.4em\relax IEEE, 2021, pp. 54--59.

\bibitem{kotaru2020multiple}
P.~Kotaru and K.~Sreenath, ``Multiple quadrotors carrying a flexible hose: dynamics, differential flatness and control,'' \emph{IFAC-PapersOnLine}, vol.~53, no.~2, pp. 8832--8839, 2020.

\bibitem{d2021catenary}
D.~S. D’antonio, G.~A. Cardona, and D.~Saldana, ``The catenary robot: Design and control of a cable propelled by two quadrotors,'' \emph{IEEE Robotics and Automation Letters}, vol.~6, no.~2, pp. 3857--3863, 2021.

\bibitem{abiko2017obstacle}
S.~Abiko, A.~Kuno, S.~Narasaki, A.~Oosedo, S.~Kokubun, and M.~Uchiyama, ``Obstacle avoidance flight and shape estimation using catenary curve for manipulation of a cable hanged by aerial robots,'' in \emph{2017 IEEE International Conference on Robotics and Biomimetics (ROBIO)}.\hskip 1em plus 0.5em minus 0.4em\relax IEEE, 2017, pp. 2099--2104.

\bibitem{estevez2022hybrid}
J.~Estevez, J.~M. Lopez-Guede, G.~Garate, and M.~Gra{\~n}a, ``Hybrid modeling of deformable linear objects for their cooperative transportation by teams of quadrotors,'' \emph{Applied Sciences}, vol.~12, no.~10, p. 5253, 2022.

\bibitem{smolentsev2023shape}
L.~Smolentsev, A.~Krupa, and F.~Chaumette, ``Shape visual servoing of a tether cable from parabolic features,'' in \emph{2023 IEEE International Conference on Robotics and Automation (ICRA)}.\hskip 1em plus 0.5em minus 0.4em\relax IEEE, 2023, pp. 734--740.

\bibitem{smolentsev2024shape}
------, ``Shape visual servoing of a cable suspended between two drones,'' \emph{IEEE Robotics and Automation Letters}, 2024.

\bibitem{gutsa2024wasted}
T.~Gutsa, C.~Trois, R.~de~Vries, and T.~Mani, ``Wasted shores: Using drones to monitor the spatio-temporal evolution of debris accumulation hotspots on south africa's umgeni river,'' \emph{Science of the Total Environment}, vol. 955, p. 176791, 2024.

\bibitem{gallitelli2024monitoring}
L.~Gallitelli, P.~Girard, U.~Andriolo, M.~Liro, G.~Suaria, C.~Martin, A.~Lusher, K.~Hancke, M.~Blettler, O.~Garcia-Garin \emph{et~al.}, ``Monitoring macroplastics in aquatic and terrestrial ecosystems: Expert survey reveals visual and drone-based census as most effective techniques,'' \emph{Science of The Total Environment}, vol. 955, p. 176528, 2024.

\bibitem{meng2020survey}
X.~Meng, Y.~He, and J.~Han, ``Survey on aerial manipulator: System, modeling, and control,'' \emph{Robotica}, vol.~38, no.~7, pp. 1288--1317, 2020.

\bibitem{ollero2021past}
A.~Ollero, M.~Tognon, A.~Suarez, D.~Lee, and A.~Franchi, ``Past, present, and future of aerial robotic manipulators,'' \emph{IEEE Transactions on Robotics}, vol.~38, no.~1, pp. 626--645, 2021.

\bibitem{richter2016polynomial}
C.~Richter, A.~Bry, and N.~Roy, ``Polynomial trajectory planning for aggressive quadrotor flight in dense indoor environments,'' in \emph{Robotics Research: The 16th International Symposium ISRR}.\hskip 1em plus 0.5em minus 0.4em\relax Springer, 2016, pp. 649--666.

\bibitem{Andersson2019}
J.~A.~E. Andersson, J.~Gillis, G.~Horn, J.~B. Rawlings, and M.~Diehl, ``{CasADi} -- {A} software framework for nonlinear optimization and optimal control,'' \emph{Mathematical Programming Computation}, vol.~11, no.~1, pp. 1--36, 2019.

\end{thebibliography}






\end{document}